\documentclass[lettersize,journal]{IEEEtran}
\usepackage{amsmath,amsfonts}
\usepackage{amssymb}
\usepackage{algorithmic}
\usepackage{algorithm}
\usepackage{array}
 \usepackage{subfigure}
\usepackage{textcomp}
\usepackage{stfloats}
\usepackage{url}
\usepackage{pifont}
\usepackage{verbatim}
\usepackage{graphicx}
\usepackage{booktabs}
\usepackage{multirow}
\usepackage{color,xcolor}
\usepackage{cite}
\usepackage[colorlinks,linkcolor=blue]{hyperref}
\usepackage{url}
\begin{document}

\title{Object Detection in Hyperspectral Image via Unified Spectral-Spatial Feature Aggregation}



\author{Xiao He, 
	Chang Tang, ~\IEEEmembership{Senior Member,~IEEE,}
	Xinwang Liu, ~\IEEEmembership{Senior Member,~IEEE,}\\
	Wei Zhang,
    Kun Sun, ~\IEEEmembership{Member,~IEEE,}
    Jiangfeng Xu
    
	\IEEEcompsocitemizethanks{
  \IEEEcompsocthanksitem This work was supported in part by the National Natural Science Foundation of China (No. 62076228 and 62176242), and in part by the Natural Natural Science Foundation of Shandong Province (No. ZR2021LZH001)
 \IEEEcompsocthanksitem X. He, C. Tang and K. Sun are with the school of computer, China University of Geosciences, Wuhan, China. E-mail: \{xiaoh, tangchang, sunkun\}@cug.edu.cn.
	\IEEEcompsocthanksitem X. Liu is with the school of computer, National University of Defense Technology, Changsha 410073, China. 
	E-mail: xinwangliu@nudt.edu.cn.
	\IEEEcompsocthanksitem W. Zhang is with Shandong Provincial Key Laboratory of Computer Networks, Shandong Computer Science Center (National Supercomputing Center in Jinan), Qilu University of Technology (Shandong Academy of Sciences), Jinan 250000, China. E-mail:  wzhang@qlu.edu.cn.
        \IEEEcompsocthanksitem J. Xu is with the Hexagon AB.
	E-mail: jiangfeng.xu@hexagon.com.
}
 
 }

\markboth{IEEE Transactions on Geoscience and Remote Sensing,~Vol.~14, August~2023}%
{Shell \MakeLowercase{\textit{et al.}}: A Sample Article Using IEEEtran.cls for IEEE Journals}


\maketitle

\begin{abstract}
Deep learning-based hyperspectral image (HSI) classification and object detection techniques have gained significant attention due to their vital role in image content analysis, interpretation, and broader HSI applications. However, current hyperspectral object detection approaches predominantly emphasize spectral or spatial information, overlooking the valuable complementary relationship between these two aspects. In this study, we present a novel \textbf{S}pectral-\textbf{S}patial \textbf{A}ggregation (S2ADet) object detector that effectively harnesses the rich spectral and spatial complementary information inherent in the hyperspectral image. S2ADet comprises a hyperspectral information decoupling (HID) module, a two-stream feature extraction network, and a one-stage detection head. The HID module processes hyperspectral data by aggregating spectral and spatial information via band selection and principal components analysis, consequently reducing redundancy. Based on the acquired spectral and spatial aggregation information, we propose a feature aggregation two-stream network for interacting spectral-spatial features. Furthermore, to address the limitations of existing databases, we annotate an extensive dataset, designated as HOD3K, containing 3,242 hyperspectral images captured across diverse real-world scenes and encompassing three object classes. These images possess a resolution of 512$\times$256 pixels and cover 16 bands ranging from 470 nm to 620 nm. Comprehensive experiments on two datasets demonstrate that S2ADet surpasses existing state-of-the-art methods, achieving robust and reliable results. The demo code and dataset of this work are publicly available at \url{https://github.com/hexiao-cs/S2ADet}.

\end{abstract}


\section{Introduction}

\IEEEPARstart{O}{bject} detection is a crucial task that involves identifying objects belonging to specific classes within images. It has a wide range of applications in various domains, including remote sensing~\cite{xia2018dota}, autonomous driving~\cite{zhao2022scene}, and medical analysis~\cite{zhang2021dynamic}. Over the past few decades, significant efforts have been made to improve object detection performance for RGB images~\cite{zhao2019object, cai2022mask, zhang2021learning}. However, hyperspectral data offer several advantages over RGB data, which can capture more intrinsic properties of object materials, enabling fine-grained object detection~\cite{miao2019net,mou2020nonlocal,hou2021hyperspectral,hong2021spectralformer}. Hence, hyperspectral detection has garnered substantial attention in the community.

\begin{figure*}[t]
				\centering
	\begin{tabular}{cc}
			\includegraphics[width = 0.9\linewidth]{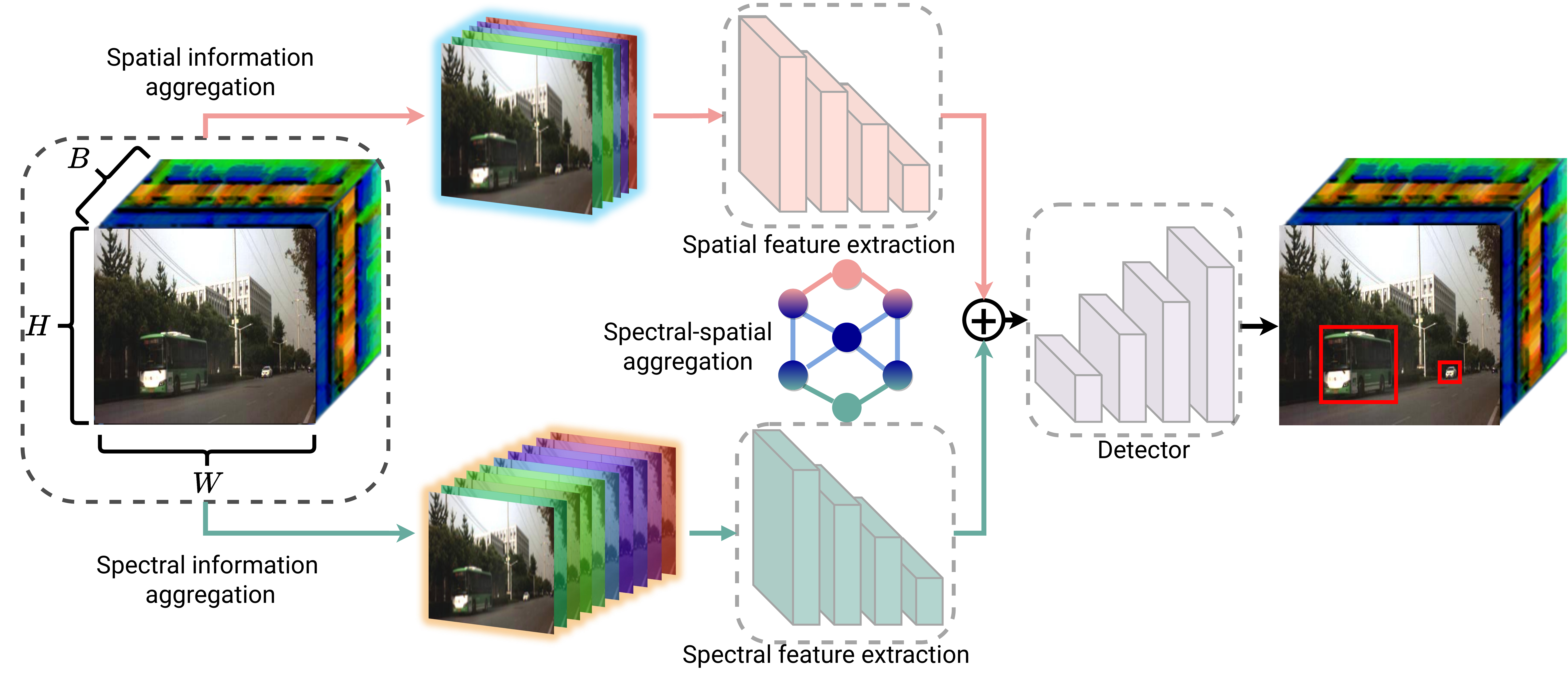}
		\end{tabular}
	\caption{Illumination of the S2ADet. The S2ADet is purposely designed to handle object detection in hyperspectral data.}
	\label{img:start}
\end{figure*}

Hyperspectral detection exhibits a pivotal role in remote sensing applications by identifying specific materials or objects through spectral features. Recent research by Rui et al. \cite{rui2021learning}, Chang et al. \cite{chang2020effective}, and Dong et al. \cite{dong2021asymmetric} primarily rely on either pixel-level spectral information or correlation information between neighboring pixels. These approaches are divided into two main categories: statistical distribution and subspace models. Statistical distribution models, as described by Shang et al. \cite{shang2020target} and Chang et al. \cite{chang2021hyperspectral}, assume that the background of the image conforms to a multivariate normal distribution. These models detect anomalous pixels by comparing them to the center of the background distribution, enabling the detection of potential objects within the anomalous pixel region. On the other hand, subspace models \cite{shang2020target,sun2021constrained,liu2016tensor} typically involve designing a linear filter that minimizes the output energy while satisfying a constraint equation. This approach allows for identifying subspaces containing the signal of interest, leading to accurate hyperspectral target detection.

Object detection in hyperspectral images is a broader concept that does not require knowledge of the spectral characteristics of the target of interest. Recently, Yan et al. \cite{yan2021object} introduced a pioneering deep learning method for leveraging the spatial attributes of hyperspectral data. Moreover, they created a HOD-1 dataset for object detection in hyperspectral images, comprising 454 images with annotated bounding boxes that tightly encircle the edges of the objects. To the best of our knowledge, HOD-1 is the first object detection dataset for hyperspectral images.

However, most current hyperspectral detection methods primarily rely on the spectral information of specific image bands to identify materials, neglecting the critical semantic information contained in the spatial dimension. In contrast, existing deep learning-based object detection methods in hyperspectral images focus on spatial information and do not consider the rich spectral semantic information available. Furthermore, the hyperspectral image in neighboring bands are highly similar and contain a large amount of redundant information between neighboring bands, and simply feeding them into the feature extractor can significantly hinder the detection performance.

To address this challenge, we propose S2ADet, a novel object detector with a two-stream spectral-spatial feature aggregation approach to detect objects in hyperspectral data. S2ADet leverages complementary spatial and semantic information to learn better semantic features of objects. The detector comprises a hyperspectral information decoupling (HID) module, a two-stream feature extraction network, a spectral-spatial aggregation (SSA) module, and a one-stage detection head. First, to address the issue of hyperspectral information redundancy and aggregate the spectral and spatial information, the HID module uses principal component analysis (PCA) to obtain \textbf{S}p\textbf{E}tral (SE) information by aggregating spectral dimension data, as shown in Fig~\ref{img:start}. Furthermore, we employ a band selection algorithm to select representative bands that retain high-quality spatial information, generating \textbf{S}p\textbf{A}tial (SA) information as another input to the network. Next, the spectral and spatial semantic information is extracted separately using a two-stream network. An SSA module is embedded in the network to facilitate the spatial interaction and aggregation of the spectrum between the two network streams. Finally, the aggregated features are input to a one-stage object detector to obtain fine-grained object bounding boxes and classification results.

\begin{figure*}[t] \centering
	\includegraphics[width = 1\linewidth]{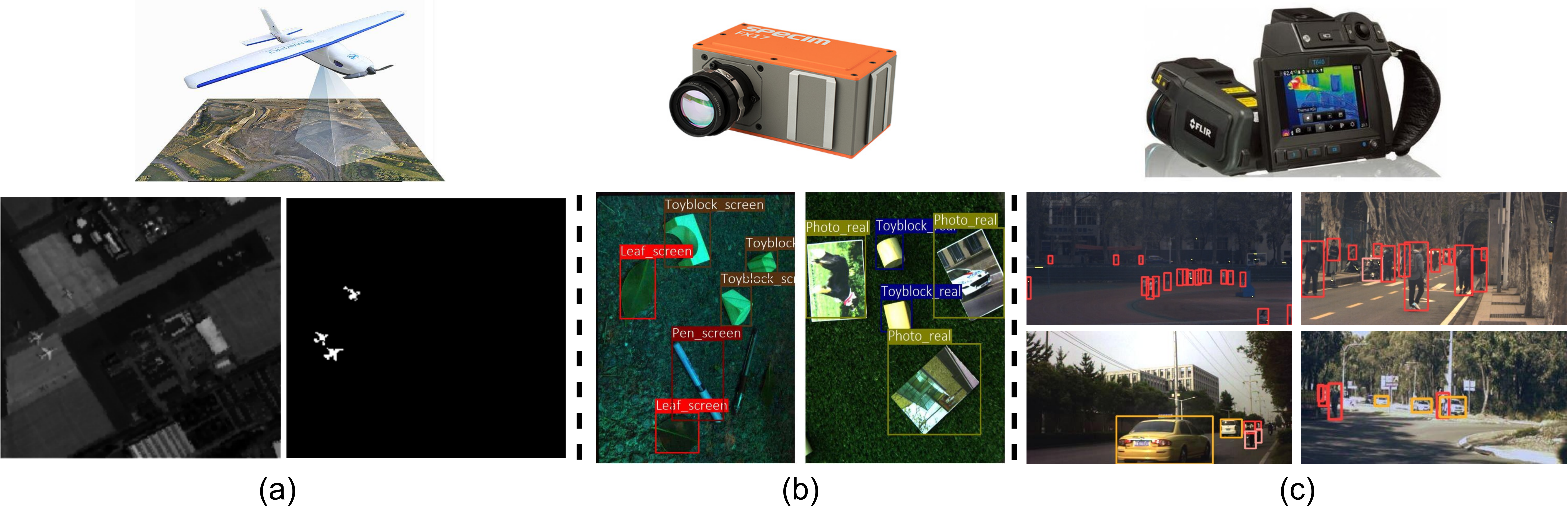}
	\caption{Comparison of existing benchmarks and datasets. (a) Hyperspectral target detection dataset. (San Diego as an example) where targets are represented as pixel-level categories and annotations. (b) Example of HOD-1 dataset. The annotations are boxes and categories immediately adjacent to the edges of the objects. Its scene is constructed by posing objects. (c) Example of our HOD3K dataset with diverse scenes.} \label{img:contrast}
\end{figure*}

Furthermore, current hyperspectral datasets suffer from limited background distribution and small dataset sizes, as observed in HYDICE, San Diego~\cite{neuenschwander1998mapping}, and Cuprite~\cite{resmini1997mineral}, as illustrated in Fig~\ref{img:contrast}. To overcome this shortcoming and promote the development of object detection in hyperspectral images, we have annotated a dataset, namely HOD3K, comprising 3,242 hyperspectral images of urban roads, campuses, and residential areas, containing 15,149 objects classified into three categories. HOD3K is an extensive hyperspectral dataset that provides a valuable resource for researchers to enhance the effectiveness of object detection methods.

The main contributions of this paper are summarized as follows:
\begin{itemize}

	\item We propose a novel detector dedicated to hyperspectral object detection, namely S2ADet. It contains a HID module, a two-stream network, and a SSA module. The detector achieves the most superior performance and provides uniform input to the network during training and inference.
	
	\item We design the HID module to decouple hyperspectral images, thereby reducing redundancy and aggregating the spectral and spatial information.

    \item We designed a two-stream network and facilitated the aggregation of spectral and spatial information by the SSA module, thus extracting refined features.
   
	\item We annotated a comprehensive large-scale object detection dataset in hyperspectral images. It can provide profound insights for developing detection.
 
\end{itemize}

The remainder of this paper is structured as follows. In Section \ref{sec:sec2}, we provide a brief review of the most related work. Section \ref{sec:sec3} elaborates on the specifics of the proposed method. Next, in Section \ref{sec:sec4}, we provide a detailed description of the proposed dataset. Section \ref{sec:sec5} comprises a series of experiments, discussions, and model analyses. Finally, we conclude this paper in Section \ref{sec:sec6}.

\section{Related Work}\label{sec:sec2}

\subsection{General Object Detection}

Object detection aims to perform edge regression and class classification on detected objects, and it has found numerous applications, such as in autonomous driving~\cite{sun2020scalability} and anomaly detection~\cite{zavrtanik2021draem}. Object detection methods can be categorized based on the number of stages involved in the detection head, which are two-stage methods~\cite{girshick2015fast,he2017mask} and one-stage methods~\cite{zhang2021vit,liu2016ssd}. Two-stage object detection methods have been rapidly evolving since the introduction of RCNN by Girshick et al.\cite{girshick2014rich}, and a series of detectors based on the Faster R-CNN\cite{ren2015faster} method have been proposed and achieved notable results. Although the two-stage detector has higher accuracy than the one-stage detector, the region proposal step in the detection process reduces the speed of object detection~\cite{liu2020deep}.

Several highly efficient one-stage object detectors have emerged to facilitate scene applications requiring high detection speed. One-stage detectors have found widespread use in industrial scenarios~\cite{redmon2017yolo9000}. Feng et al.\cite{feng2021tood} have proposed TOOD to address the problem of conflicting classification and regression tasks by aligning them into unified parallel tasks. In recent years, anchor-free approaches have received considerable attention, and FCOS\cite{tian2019fcos} has proposed replacing the anchored frames with anchored points for object detection. To enable end-to-end object detection, Carion et al.~\cite{carion2020end} have proposed DETR, which uses learnable queries to detect objects.

\subsection{Intelligent Interpretation of HSI}

With the advent of sensor technology, hyperspectral imaging has emerged as a key area of interest for numerous researchers~\cite{hang2020classification,bai2020class,liu2020few,zhu2019binary}. Hyperspectral data capture objects or scenes across a broad range of wavelengths in the electromagnetic spectrum. Compared to conventional color images, hyperspectral data can provide more detailed and accurate information~\cite{sellami2022deep,wang2019hyperspectral,li2023lightweight}. Intuitively, RGB images are composed of three channels. In contrast, hyperspectral images usually consist of numerous channels or bands ($B$). Images containing more than ten bands are generally classified, while images with more than three but fewer than ten bands are considered multispectral images~\cite{li2021mapping}. Compared with RGB images, hyperspectral images contain many bands, with each pixel containing a representative feature of the captured substance. Its finds widespread use in applications such as crop monitoring~\cite{farmonov2023crop}, target tracking~\cite{liu2021anchor}, military surveillance~\cite{koz2019ground}, and many others.

\begin{table*}[t]
	\caption{Comparison of existing datasets. Where the mark is pixel-level classification and annotation, the resolution is the maximum resolution of the video/image contained in the dataset, and HBB denotes the horizontal bounding box. Place indicates a scene constructed by placing objects, which generally does not appear in real life, as shown in Fig.~\ref{img:contrast}(b).}
	\centering
	\begin{tabular}{@{}ccccccccccc@{}}
		\toprule[1.5pt]
		\multicolumn{1}{c|}{{Hyperspectral  datasets}} &{Scene}  & Categories &Resolution& Bands  & \#Images& Annotations & {{Avg.  \#labes/images}}& Labeling method   &Year	\\			\midrule
		\multicolumn{1}{c|}{Muufl Gulfport}&{aerial}&1& 337 $\times$ 325 & 72 & 1 &-&-&mark&-  \\
		\multicolumn{1}{c|}{Nuance Cri} & aerial&1& 400 $\times$ 400& 46 & 1 &-&-&mark&-   \\
		
		\multicolumn{1}{c|}{San Diego} & aerial&1& 400 $\times$ 400 & 224 & 1 &3&3&mark&1998  \\
		\multicolumn{1}{c|}{HS-ISD~\cite{fang2023hyperspectral}} & aerial&1& 301 $\times$ 298 & 48& 56 &1085&19.37&mark&2023   \\
		\multicolumn{1}{c|}{HOD-1~\cite{yan2021object}} & place&8& 467 $\times$ 336& 96 & 454 &1657&{3.65}&HBB&2021   \\
		\multicolumn{1}{c|}{HOD3K} & natural &3&512 $\times$ 256& 16  & 3242 &15149&4.37&HBB&2023   \\
		\bottomrule[1.5pt]
	\end{tabular}
	\label{tab:dataset}
\end{table*}

\textbf{Dataset.}
In recent years, several hyperspectral datasets have been proposed for target detection, such as Muufl Gulfport, Nuance Cri, and San Diego, mainly consisting of a single remote sensing image~\cite{vali2020deep,hong2020learning,fang2023towards}. Fang et al.\cite{fang2023hyperspectral} annotated an instance segmentation dataset consisting of 56 hyperspectral images to address the gap in hyperspectral data. Yan et al.\cite{yan2021object} created a dataset for camouflage object detection by constructing scenes with posed objects, leveraging the ability to analyze the physics using hyperspectral spectral dimensional feature information. However, these datasets are restricted in size and scene diversity~\cite{wu2023querying,li2019deep}. To address this issue, we propose a new hyperspectral object detection dataset, namely HOD3K, with annotations in practical application to meet the requirements of various practical applications. Example images from the HOD3K dataset and existing representative datasets (San Diego and HOD-1) are shown in Fig.\ref{img:contrast}. As shown in Table~\ref{tab:dataset}, our proposed dataset is distinct from others in that it contains ten times more images and annotations than the most prominent existing hyperspectral dataset (HOD-1) and includes diverse scenes such as campuses, roads, and living areas. The HOD3K dataset is explicitly designed for object detection in hyperspectral images, making it a valuable resource for studying detection in large-scale scenes. A detailed comparison of the HOD3K dataset with other related datasets for object detection is presented in Table~\ref{tab:dataset}.

\begin{table}[t]
	\centering
	\caption{Distribution of annotated bounding boxes by category in the HOD3K dataset.}
		\begin{tabular}{c|c|ccc|c}
			\toprule[1.5pt]
			Dataset&Annotation & People& Car & Bike&All  \\
			\midrule	
			\multirow{2}{*}{HOD3K}&Number &12,144& 817& 2,188 & 15,149 \\ 
			\cmidrule(l){2-6} 
					{} &Ratio&80.2\%& 14.4\%& 5.4\% & 100\% \\ 
			\bottomrule[1.5pt]
		\end{tabular}
	\label{tab:class}
\end{table}

\textbf{Hyperspectral Detection.} Hyperspectral imaging captures rich image feature information employing dozens of contiguous spectral bands. However, the information redundancy between adjacent spectral bands increases the computational cost, making it challenging to leverage hyperspectral images on a large scale, especially with equipment limitations. To eliminate redundant information in hyperspectral images, Wang et al.\cite{wang2020hyperspectral} mapped the bands into subspaces and selected a combination of bands with more information, less correlation, and better category differentiability. Cheng et al.\cite{chen2021hyperspectral} utilized principal component analysis to extract the representative features of hyperspectral data. However, due to the limitation of sensors, the available hyperspectral image dataset could be more robust. Thus, research on HSI intelligence interpretation has mainly focused on image classification and semantic segmentation, with less research on object detection.

Existing methods for hyperspectral target detection mainly rely on pixel-level spectral information or simple correlation information between neighboring pixels. For instance, Chang et al.\cite{chang2020effective} proposed a statistical distribution model that detects anomalous pixels by comparing them with the center of the background distribution. This method assumes that the background follows a multivariate normal distribution and can detect possible objects in the region of anomalous pixels. Similarly, the subspace model involves designing a linear filter that minimizes the output energy while satisfying the constraint equations\cite{chang2021orthogonal}. 

To leverage in-depth information from hyperspectral images using deep learning, Yu et al.~\cite{yu2017convolutional} proposed a CNN-based object detector that integrates hyperspectral images into existing object detection models using 3D convolutional layers. However, The aforementioned methods ignore the rich contextual semantic complementary information in the spectral and spatial dimension information. To address these limitations, we propose a hyperspectral information decoupling (HID) module that decouples hyperspectral data from a spectral and spatial perspective and a two-stream network to interact with spectral and spatial dimensional information. Thus, we aim to aggregate the rich contextual semantic information in the spatial dimension with the spectral dimension information.
\begin{figure*}[t]
	\centering
	\begin{tabular}{c}
		\includegraphics[width = 1\linewidth]{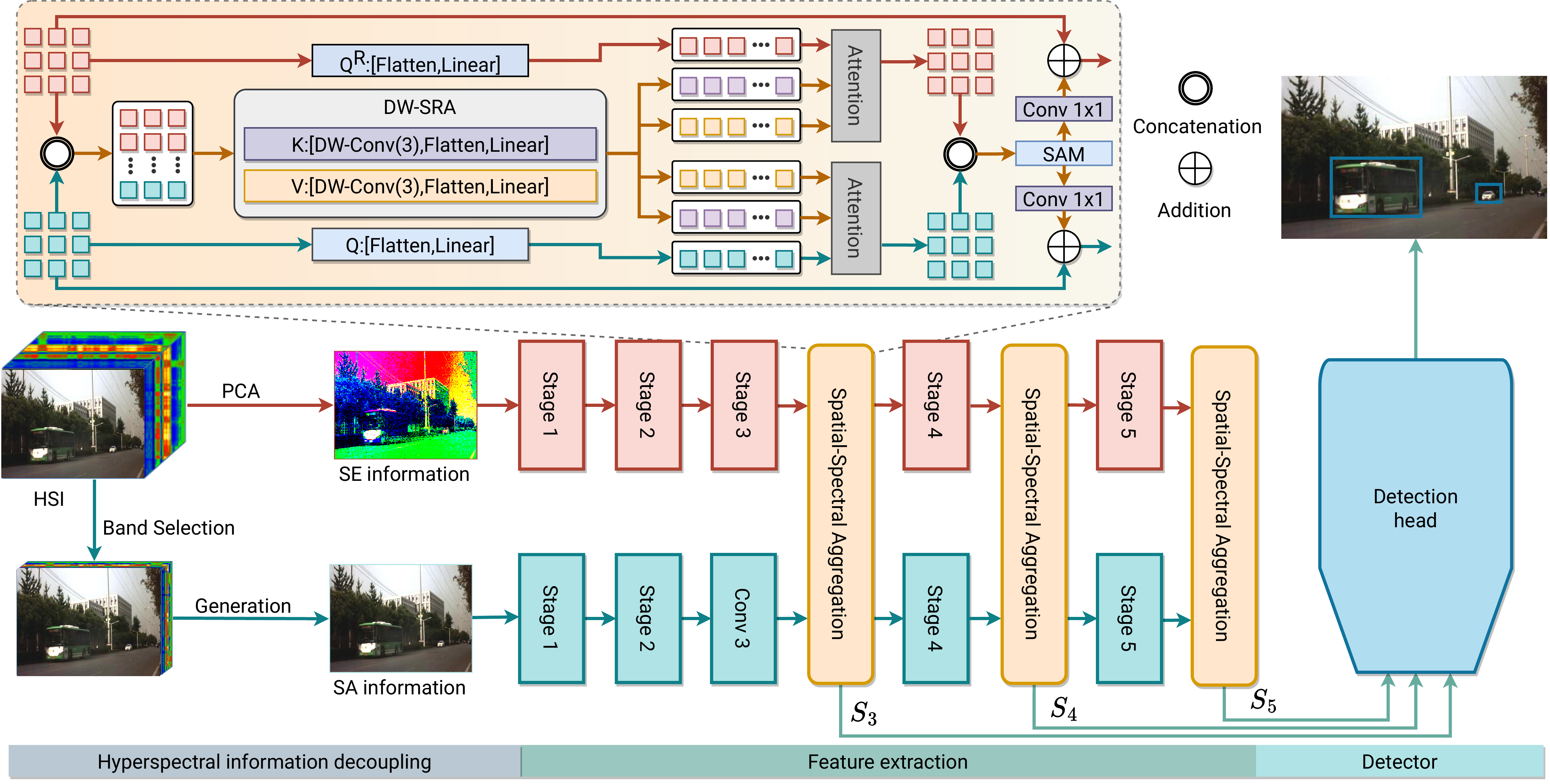}
	\end{tabular}
	\caption{The structure of S2ADet. It contains a hyperspectral information decoupling (HID) module, a two-stream network, a spectral-spatial aggregation (SSA) module, and a one-stage detection head. The hyperspectral information decoupling module is used to remove redundant information. We designed a two-stream feature extraction network embedded with SSA modules for interacting with the spectral and spatial dimensional features of hyperspectral images.}

	\label{img:overview}
\end{figure*}
\section{HOD3K Dateset}\label{sec:sec3}

Hyperspectral imaging has been extensively utilized in remote sensing applications for many years, typically satellite sensors. However, recent advancements in sensor technology have facilitated the acquisition of hyperspectral data from natural scenes to a considerable extent. The hyperspectral object detection dataset, HOD-1, introduced by Yan et al.~\cite{yan2021object}, has played a crucial role in detecting camouflaged objects by providing 454 hyperspectral images captured in 96 spectral bands, featuring 1657 annotations. However, the limitations of the dataset size and single scene availability pose significant challenges to developing large-scale hyperspectral object detection datasets for natural scenes.

To promote the development of hyperspectral data, we have developed the HOD3K dataset, inspired by~\cite{xiong2020material}. We have transformed hyperspectral images into pseudo-color images and annotated them using Labelme~\cite{torralba2010labelme}. The HOD3K dataset includes 3242 hyperspectral images of natural scenes acquired using a XIMEA snapshot VIS camera. Each hyperspectral image contains 512 $\times$ 256 pixels with 16 bands arranged in a mosaic pattern spanning the range of 470 nm to 620 nm. The dataset includes various scenarios, such as urban roads and campuses, presenting challenges like occlusion, small objects, and scale changes. We have provided finely annotated bounding boxes for all objects belonging to the three categories: people, cars, and bikes, resulting in a total of 15,149 objects, with an average of 4.67 objects per image. Specifically, the dataset comprises 12,144 people, 817 cars, and 2,188 bikes. Table~\ref{tab:class} shows the distribution of the number of tags in each category. The availability of the dataset provides researchers with a more extensive and diverse set of natural scene images and presents new challenges for developing and evaluating object detection algorithms. Moreover, we use pseudo-color images for refined annotation, allowing researchers to develop and test new techniques to advance the field of hyperspectral detection further.

\section{Proposed Method}\label{sec:sec4}
In this section, we present an overview of the proposed method. Subsequently, we shall expound upon the intricate workings of the hyperspectral information decoupling module, spectral-spatial aggregation module, the two-stream network, and the detector. Finally, We elucidate the construction of the loss function.

\subsection{Framework Overview}

Fig.~\ref{img:overview} depicts the overall architecture of our novel object detection detector, named S2ADet, which leverages the rich spectral and spatial information in hyperspectral images. To this end, we employ a feature pyramid network (FPN)~\cite{lin2017feature} and detector head inspired by the one-stage object detector~\cite{redmon2017yolo9000}. For the sake of brevity, we omit the FPN and detector head from the figure.

First, the problem of redundant hyperspectral information is solved by band selection and principal component analysis, which decouples hyperspectral information into spectral (SE) information and spatial (SA) information. Then, we introduce a two-stream backbone network that utilizes a recursive structure to extract the features of the aggregated images. The network comprises five stages, with the features of the SA information in each stage denoted as ${a}^1, {a}^2, \cdots, {a}^5$, and the features of the SE information in each stage denoted as ${e}^1, {e}^2, \cdots, {e}^5$. The two-stream backbone is illustrated at the bottom of Fig.~\ref{img:overview}, and the structure of the first three layers of the feature extraction network is described as follows:
\begin{equation}
	{a}^{i+1} = \textrm{Stage}_i\left({a}^{i} \right), {e}^{i+1} = \textrm{Stage}_i\left({e}^{i} \right),
\end{equation}
where $\textrm{Stage}_i$ is the feature extraction of the $i$-th stage. And then, to effectively integrate both the spectral and spatial information from the hyperspectral image, we introduce a spectral-spatial aggregation module (SSA) to facilitate interactions between spectral and spatial features. The SSA module is inserted into the last three layers of a two-stream network and operates on a tensor containing both spectral and spatial features. The module outputs two tensors, one for aggregated spectral information and another for aggregated spatial information. Specifically, we use a stage-wise approach, where the aggregated information is added to the feature maps in each stage. The equation represents it:
 \begin{equation}
	\begin{gathered}
		{\left[\begin{array}{l}
				{a}^{i}_{\mathcal{T}} \\
				{e}^{i}_{\mathcal{T}}
			\end{array}\right]=\textrm{SSA}\left(\left[\begin{array}{l}
				{a}^{i}  \\
				{e}^{i} 
			\end{array}\right]\right),} \\
				{\left[\begin{array}{l}
				{a}^{i+1} \\
{e}^{i+1}
			\end{array}\right]= \textrm{Stage}_i\left(\left[\begin{array}{l}
				{a}^{i} \\
				{e}^{i}
			\end{array}\right]+\left[\begin{array}{l}
				{a}^{i} _{\mathcal{T}} \\
				{a}^{i} _{\mathcal{T}}
			\end{array}\right]\right), }\\
	\end{gathered}
\end{equation}
where ${a}^{i}$ and ${e}^{i}$ denote the spectral and spatial feature maps at stage $i$, respectively. The output of the SSA module, denoted as ${a}^{i}_{\mathcal{T}}$ and ${e}^{i}_{\mathcal{T}}$, are concatenated to the input feature maps before being passed to the next stage. Then, The resulting features are subsequently fed into the detection head. This operation is mathematically represented as:
\begin{equation}
	{S}_i=\textrm{Add}\left(\widehat{{a}}^{i},\widehat{{e}}^{i}\right),
\end{equation}
where ${S}_3$, ${S}_4$, and ${S}_5$ are the resulting features utilized as input to the object detector module. After passing through the object detector, the final detection result is obtained.

\subsection{Hyperspectral Information Decoupling}

Hyperspectral imaging instruments demonstrate variability in capturing spectral bands, leading to high information redundancy within hyperspectral images. Hence, the direct processing of hyperspectral images using the detector ($H$) limits the detector's performance. To overcome this challenge, we decouple the hyperspectral image from spectral and spatial perspectives and process the raw data by a hyperspectral information decoupling (HID) module, as shown in Fig.~\ref{img:overview}. This process can be expressed as follows:
\begin{equation}
	{a} = \textrm{SpatialFilter}\left(H \right), {e} = \textrm{SpectralFilter}\left(H \right),
\end{equation}

\textbf{Spatial Aggregated Information.}
	Hyperspectral imaging is characterized by a high correlation between adjacent spectral bands, leading to significant spatial data redundancy. To reduce the redundancy between hyperspectral bands, we introduce a parameter-free method known as optimal neighborhood reconstruction (ONR)~\cite{wang2020hyperspectral} is employed to select the most informative spectral bands from the hyperspectral data. This band selection step creates a spatial dimension aggregation image that preserves the most relevant spectral information. Furthermore, to further refine the spatial information, we employ a technique known as color mapping to generate SA information (${a}$).

\textbf{Spectral Aggregated Information.}
The hyperspectral image contains more than ten times the number of bands in natural images, leading to a considerable reduction in the object detection speed. To address this challenge posed by the redundancy of spectral data in hyperspectral images, we have utilized principal component analysis~\cite{abdi2010principal} for compressing these images based on their spectral dimensions. Through dimensionality reduction, we obtained SE information (${e}$).

\subsection{Spectral-Spatial Aggregation module}
After creating the hyperspectral information decoupling module, the SE and SA information can be acquired. In order to extract the features of the interacted images in both spectral and spatial dimensions, we have designed a two-stream network. As shown in the top of Fig.~\ref{img:overview}, the network performs feature extraction from the SE and SA information, combining attention operations to form the spectral-spatial aggregation (SSA) module.

The input ${a}^{i} \in \mathbb{R}^{n \times d}$ is the feature input from the SA information, while ${e}^{i} \in \mathbb{R}^{n \times d}$ denotes the SE information. Here, $n$ denotes the number of patches, which is equal to the product of the height ($h$) and width ($w$) of the image, and $d$ denotes the dimension of the features. Firstly, To achieve contextual information aggregation of spectral-spatial image features, we concatenate together two image features as follows:
\begin{equation}
	 {f}^{i}=\textrm{Concat}\left({a}^{i}, {e}^{i}\right).
\end{equation}
Then, ${f}^{i}$ as the common ${f}^{i} W_K \in \mathbb{R}^{n \times d_k \times 2}$ and ${f}^{i} W_V \in \mathbb{R}^{n \times d_v \times 2}$, ${a}^{i}$ as ${a}^{i} W_Q \in \mathbb{R}^{n \times d_v}$, and ${e}^{i}$ as ${e}^{i} W_Q \in \mathbb{R}^{n \times d_v}$, and feed into the separate transformer modules. Since calculating attention directly through hyperspectral features would entail high computational effort, we use spatial reduction attention~\cite{guo2022cmt} to reduce the dimension of the key and value.

\begin{equation}
	\begin{aligned}
		&{f}^{i} \tilde{W}_K=\textrm{DWConv}({f}^{i} W_K) \in \mathbb{R}^{\frac{n}{r^2} \times d_k \times 2},\\
		&{f}^{i} \tilde{W}_V=\textrm{DWConv}({f}^{i} W_V) \in \mathbb{R}^{\frac{n}{r^2} \times d_k \times 2},
	\end{aligned}
\end{equation}
where $r$ is the spatial reduction rate of the SRA. Then, ${a}^{i} W_Q, {e}^{i} W_Q, {f}^{i} \tilde{W}_K, {f}^{i} \tilde{W}_V$  are put into the attention block separately to calculate.
\begin{equation}
	\begin{aligned}		
		&	\textrm{Attn}_1({a}^{i} W_Q, {f}^{i} \tilde{W}_K, {f}^{i} \tilde{W}_V)=\textrm{Softmax}\left(\frac{att({a}^{i}, {f}^{i})}{\sqrt{d_k}}\right) {f}^{i} \tilde{W}_V,\\
		&	\textrm{Attn}_2({e}^{i} W_Q, {f}^{i} \tilde{W}_K, {f}^{i} \tilde{W}_V)=\textrm{Softmax}\left(\frac{att({e}^{i}, {f}^{i})}{\sqrt{d_k}}\right) {f}^{i} \tilde{W}_V,\\
	\end{aligned}
\end{equation}
where $att(x, y)=\left(x W_Q\right)\left(y W_K\right)^T $. Then the FFN layer is sent to add nonlinear transformations. The overall architecture after transformers is expressed formally as:
\begin{equation}
	\begin{gathered}
		{\left[\begin{array}{l}
				\widetilde{{a}}^{i} \\
				\widetilde{{e}}^{i}
			\end{array}\right]=\left[\begin{array}{l}
				{a}^{i} \\
				{e}^{i}
			\end{array}\right]+\textrm{Att}\left(\left[\begin{array}{l}
				{a}^{i} \\
				{e}^{i}
			\end{array}\right]\right),} \\
		{\left[\begin{array}{l}
				\overline{{a}}^{i} \\
				\overline{{e}}^{i}
			\end{array}\right]=\left[\begin{array}{l}
				\widetilde{{a}}^{i} \\
				\widetilde{{e}}^{i}
			\end{array}\right]+\textrm{FFN}\left(\left[\begin{array}{l}
				\widetilde{{a}}^{i} \\
				\widetilde{{e}}^{i}
			\end{array}\right]\right) .}
	\end{gathered}
\end{equation}
We coupled interactive hyperspectral information by leveraging transformer global and dynamic modeling capabilities. Hence, this process ultimately aggregates contextual information based on spectral-spatial dimensions.

\begin{figure}[t]
	\centering
	\begin{tabular}{c}
		\includegraphics[width = 1\linewidth]{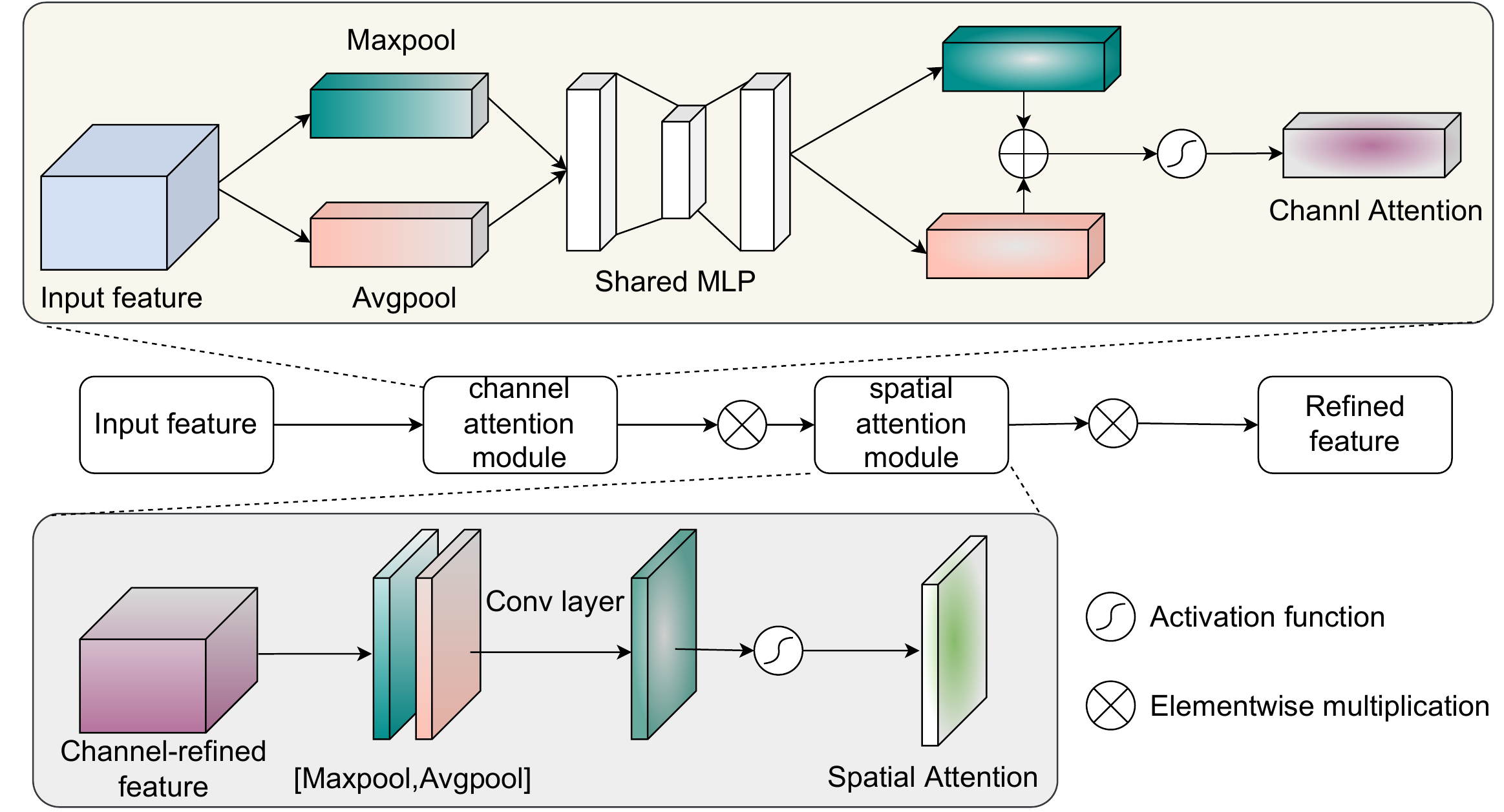}
	\end{tabular}
	\caption{Illustration of SAM module. The SAM module comprises a channel attention module and a spatial attention module.}
	\label{img:cbam}
\end{figure}
\textbf{Spectral Attention module.}  In order to interact with the information of spectral dimensions, we further used the SAM module~\cite{woo2018cbam} to calculate the channel attention (shown in Fig.~\ref{img:cbam}), and the module can be formulated as:
 \begin{equation}
 	\begin{aligned}		
  & \widetilde{{f}}^{i}=\textrm{Concat}\left(\overline{{a}}^{i}, \overline{{e}}^{i}\right),\\
 & \widetilde{{f}}^{i}_c=\textrm{M}_c(\widetilde{{f}}^{i}) \otimes \widetilde{{f}}^{i}, \\
 & \widetilde{{f}}^{i}_s=\textrm{M}_s\left(\widetilde{{f}}^{i}_c\right) \otimes \widetilde{{f}}^{i}_c,\\
 	\end{aligned}		
\end{equation}
where $M_c(\cdot)$ is the channel attention map, $M_s(\cdot)$ is the spatial attention map, and $\otimes$ stands for elementwise multiplication. Then, we mapped them into the respective feature extraction networks by $\mathbf{1 \times 1}$ convolution, aligning the number of feature extraction network channels and continuing the extraction operation.
\begin{equation}
	\begin{aligned}		
				{a}^{i}_{\mathcal{T}}=\textrm{Conv}_{1\times 1}\left(\widetilde{{f}}^{i}_s\right),	{e}^{i}_{\mathcal{T}}=\textrm{Conv}_{1\times 1}\left(\widetilde{{f}}^{i}_s\right),
	\end{aligned}		
\end{equation}
 where $\widetilde{{f}}^{i} \in \mathbb{R}^{n \times d \times 2}$. ${a}^{i}_{\mathcal{T}}, {e}^{i}_{\mathcal{T}} \in \mathbb{R}^{n \times d}$. Through this operation, the spectral and spatial information of the hyperspectral image is aggregation.

\textbf{Loss Function.}
For the task of HSI object detection, the multitask loss function is defined as follows during training:
\begin{equation}
	L=L_{\mathrm{cls}}+L_{\mathrm{box}},
\end{equation}
where $L_{\mathrm{cls}}$ is the classification loss and $L_{\mathrm{box}}$ is the bounding-box loss. The decouples box and class prediction differ from common practice when applied to object detection. Several experimental results from previous work have proved that the loss function is the key to good object detection results. Our subsequent experiments show that this loss function applies to hyperspectral object detection.


\begin{table*}[t]
	\caption{Comparison to methods on HOD3K dataset with resolution at 512 $\times$ 256. SA denotes the image of spectral aggregated information, and SE denotes the image of spectral aggregated information.}
	\begin{center}
		\begin{tabular}{c|c|c|cccc|c|c|c|c}
			\toprule[1.5pt]
			Detectors &Backbone&Type  & people &bike & car & mAP50 & mAP   &FLOPs  & Param.& Input 	\\
			\midrule
			{Faster RCNN~\cite{ren2015faster}} &ResNet50 &two-stage& {81.8} & {94.5} &{91.7}&{89.4}&{56.9}&206.68 &41.14M& SA \\ 
			{Libra RCNN~\cite{pang2019libra}} &ResNet50 &two-stage& {83.6} & {95.0} &{90.8}&{89.8}&{56.6} &207.73 &41.40M & SA\\ 
			\midrule
			{FCOS~\cite{tian2019fcos}} & ResNet50&one-stage& {55.0} & {19.7} &{69.8}&{48.2}&{23.7}& 196.81 &31.84M & SA \\
			
			{YOLOF~\cite{chen2021you}} & ResNet50 &one-stage& {60.8} & {67.8} &{67.6}&{65.4}&28.6 &98.23 &{42.13M}&SA\\	 
		{Deformable DETR~\cite{zhu2020deformable}} & ResNet50 &one-stage& {52.8} & {56.3} &{64.9}&{58.0} &22.3&195.23&{39.82M}&SA\\	
			
			{Retinanet~\cite{lin2017focal}} &ResNet50&one-stage & {85.6} & {94.8} &{92.6}&{91.2}&{53.3}&205.69 &36.17M& SA\\ 
			{TOOD~\cite{feng2021tood}} &ResNet50 &one-stage& {85.1} & {87.0} &{89.6}&{87.2}&{55.4}&180.66&31.80M& SA\\
			{YOLOv5} & DarkNet53 &one-stage& {79.3} & {94.0} &{91.2}&{88.1}&{54.4} &\textbf{48.30} &\textbf{20.88M}&SA\\	 
			\multicolumn{1}{c|}{YOLOv5\dag} & \multicolumn{1}{c|}{DarkNet53} &\multicolumn{1}{c|}{one-stage}& {83.6} & {96.4} &{95.2}&{91.7}&{56.3}& 89.72 &35.49M& {SA + SE} \\	
			\midrule
			
			{S2ADet*} & S2ANet &one-stage&{83.8} &  {94.3} &{92.7}&{90.3}&{55.5} &169.20 & 48.64M  &SA + SA\\
			{S2ADet(Ours)} & S2ANet &one-stage& \textbf{87.2} & \textbf{97.7} &\textbf{95.3}&\textbf{93.4}&\textbf{59.8} &169.20 & 48.64M  & SA + SE \\
			\bottomrule[1.5pt]
		\end{tabular}
	\end{center}
	\label{tab:HOD3K}
\end{table*}

\begin{table*}[t]
	\centering
	\caption{Comparison to methods on HOD-1 dataset. The screen is denoted as S, and real is denoted as R. HSI is denoted as the hyperspectral image.}

	\begin{tabular}{c|c|ccccccccccc} 
		\toprule[1.5pt]
		Detectors&Backbone &toyblock S&photo S&pen S&photo R&toyblock R&pen R&leaf S&leaf R & mAP\\
		\midrule	
		{FCOS~\cite{tian2019fcos}} & ResNet50&-&-&-&-&-&-&-&-&{80.9}  \\
		{Double-Head RCNN~\cite{wu2020rethinking}} &ResNet50&-&-&-&-&-&-&-&- & {81.2}  \\
		{FoveaBox~\cite{kong2020foveabox}} & ResNet50&-&-&-&-&-&-&-&- & {80.2} \\
		{YOLOv5} & DarkNet53 &82.4&80.3&34.0&97.7&97.6&76.9&81.1&93.3&80.4 \\
		{YOLOv5} & DarkNet53 &81.6&83.0&33.3&\textbf{98.6}&97.3&77.0&79.6&94.7&80.7  \\
		{HOD-1~\cite{yan2021object}} & VGG16&-&-&-&-&-&-&-&- & {83.5}  \\
		{S2ADet (Ours)} & DarkNet53 &\textbf{83.1}&\textbf{88.3}&\textbf{51.5}&{98.4}&\textbf{98.0}&\textbf{91.2}&\textbf{83.3}&\textbf{98.7}&\textbf{86.6}  \\
		\bottomrule[1.5pt]
	\end{tabular}
	\label{tab:hsi}
\end{table*}

\section{Experiments}\label{sec:sec5}

In this section, we present our experimental settings, ablation study, extensive results with in-depth analysis, and comparisons with competing methods in detail.

\subsection{Experimental Setting}

\emph{1) Implementation Details:}
We used DarkNet-FPN as the two-stream backbone network, and the pretrained DarkNet-50 was used for initialization. We conducted comprehensive experiments on two datasets to demonstrate the effectiveness of our proposed method. All experiments trained 50 epochs. We use a mosaic of data enhancement techniques. For a fair comparison, all experiments were trained and tested on a single NVIDIA RTX 3090, using an SGD optimizer with a learning rate of 0.01. The $poly$ learning scheme is adopted so that the learning rate is adjusted as $(1 - \frac{epoch}{max\_epoch} )^{power} \times lr$, where $power = 0.9$. The threshold for non-maximum suppression (NMS) IoU in testing was 0.6. 

\emph{2) Evaluation Metrics:}
We evaluate S2ADet on the two datasets using the Average Precision (AP) evaluation metric. According to the IoU threshold (i.e., 0.50:0.95, 0.50, 0.75), AP makes a finer division of the evaluation metrics.

\subsection{Datesets and Evaluation metrics}
In our experiments, we use two hyperspectral object detection benchmark datasets. The detailed information is listed as follows.

\textbf{HOD3K.}
The HOD3K dataset consists of 16 bands and includes various natural scenes. In constructing the dataset partitioning protocol, we considered two primary factors. Firstly, we ensured that the distribution of objects within each category was consistent across all sets (training, validation, and test). Secondly, we randomly selected images from different scenes to create a ratio of 7:1:2 for the training, validation, and test set, respectively. Subsequently, we obtained segmented training, validation, and test sets. We trained the methods on the training set for all experiments and evaluated them on the test set.

\textbf{HOD-1.}
The HOD-1 dataset is a recently introduced hyperspectral object detection dataset comprising 1657 hyperspectral images, categorized into eight distinct categories, with a band count of 96. The dataset generates an object detection scene by strategically placing objects, which are then captured using a hyperspectral camera. The dataset also features a camouflaged scene, created by initially photographing the real scene with an iPadAir camera and then capturing the image on the iPadAir screen using the hyperspectral camera. Thus, both the real and camouflaged images are available for analysis.

\textbf{Evaluation metrics.}
We employ the standard metric of mean average precision (mAP) to assess the accuracy of multispectral object detection. For mAP calculations, an Intersection over Union (IoU) threshold of 0.5 is utilized to determine True Positives (TP) and False Positives (FP).

\subsection{Comparative Methods}

%

\textbf{Performance on HOD3K Dataset.} We compared our proposed S2ADet method with state-of-the-art advanced object detection methods, including Faster RCNN~\cite{ren2015faster}, FCOS~\cite{tian2019fcos}, Deformable DETR~\cite{zhu2020deformable}, YOLOF~\cite{chen2021you}, YOLOv5, TOOD~\cite{feng2021tood}, and Libra RCNN~\cite{pang2019libra}, which mainly focus on local features. In contrast, our method captures rich global semantic relational information, which enables it to outperform these six methods. The experimental results are presented in Table~\ref{tab:HOD3K}. Specifically, our method captures rich spectral and spatial semantic relationship information, which makes it superior to the aforementioned methods. 

Compared to the state-of-the-art method, our S2ADet method improves 2.9\% mAP over Faster RCNN, which currently holds the best performance in mAP. Moreover, compared to the one-stage object detection method, our method achieves a 3.5\% mAP improvement over the current best method. Additionally, when compared to the highest accuracy in each category of the compared methods, our S2ADet method achieves the highest detection accuracy across all three categories. Specifically, comparing the state-of-the-art algorithm YOLOv5\dag with mAP50, our method improves the accuracy of detecting people, bikes, and cars by 3.6\% mAP, 1.3\% mAP, and 0.1\% mAP, respectively. These results further confirm the effectiveness of our proposed method.

\textbf{Performance on HOD-1 Dataset.}   We conducted a series of comparative experiments using the HOD-1 dataset~\cite{yan2021object} to assess the efficacy of our proposed methodology rigorously. We compared four other state-of-the-art methods, specifically FCOS~\cite{tian2019fcos}, Double-Head R-CNN, FoveaBox, and HOD-1. A summary of the results can be found in Table~\ref{tab:hsi}. Notably, our proposed method surpasses the current leading technique (HOD-1) by achieving a 3.1\% mAP improvement in mean average precision (mAP). It is essential to mention that the input data size for our method is merely one-sixteenth of that of HOD-1. Nevertheless, our method still attains a higher mAP than the modified algorithm. This result underscores the effectiveness of our meticulously designed hyperspectral information decoupling module tailored explicitly for hyperspectral images.

Furthermore, our proposed method, S2ADet, achieves good results in each category, particularly on photo screen, pen screen, and pen real, with performance improvements of 5.3\% mAP, 18.2\% mAP, and 14.2\% mAP, respectively. These comprehensive experiments on two datasets demonstrate the robustness and effectiveness of our approach.

\begin{table*}[t]
	\centering
	\caption{Ablation study of different inputs in spectral and spatial aggregated information on HOD3K Dataset. The baseline is yolov5. SA is the image of spectral aggregated information, and SE is the image of spectral aggregated information. G indicates generation.}
	\begin{tabular}{@{}ccccccccc|c|c@{}}
		\toprule[1.5pt]
		\multicolumn{1}{c|}{Detectors} &\multicolumn{1}{c|}{Input}&\multicolumn{1}{c|}{Bands}& people &bike & car & mAP50 & mAP	\\ \midrule	
		\multicolumn{1}{c|}{\multirow{3}{*}{Baseline}}&\multicolumn{1}{c|}{SA w/o G}&\multicolumn{1}{c|}3& {3.69} & {71.4} &{66.0}&{47.0}&{34.7} \\ 
		\multicolumn{1}{c|}{}&\multicolumn{1}{c|}{SE} &\multicolumn{1}{c|}3& {75.0} & {67.8} &{90.8}&{77.9}&{45.2} \\
		\multicolumn{1}{c|}{} &\multicolumn{1}{c|}{SA}&\multicolumn{1}{c|}3& {79.3} & {94.0} &{91.2}&{88.1}&{54.4} \\
		\midrule	
		\multicolumn{1}{c|}{\multirow{3}{*}{S2ADet}} &\multicolumn{1}{c|}{SA w/o G + SA w/o G}& \multicolumn{1}{c|}{6} & {2.01} & {76.2} &{55.7}&{44.6}&{40.4} \\ 
		\multicolumn{1}{c|}{} &\multicolumn{1}{c|}{SA + SA} &\multicolumn{1}{c|}{6}&	 {83.3} & {91.7} &{90.1}&{88.3}&{57.2} \\ 
		\multicolumn{1}{c|}{}& \multicolumn{1}{c|}{SA + SE}&\multicolumn{1}{c|}{6}& \textbf{87.2} & \textbf{97.7} &\textbf{95.3}&\textbf{93.4}&\textbf{59.8} \\
		\bottomrule[1.5pt]
	\end{tabular}
	\label{tab:input}
\end{table*}
\begin{table}[t]
	\centering
	\caption{Ablation study of the SSA module in on HOD3K Dataset.}
		\begin{tabular}{c|c|c|c|c|c}
			\toprule[1.5pt]
			Detectors& Input& Bands & SSA & SAM & mAP \\
			\midrule	
			{Baseline} &SA& 3& $-$ & $-$ &{54.4} \\ 
			\midrule	
			{\multirow{4}[0]{*}{S2ADet}}&SA + SE& 6& $-$ & $-$ &{56.3} \\
			{}  &SA + SE & 6& $-$ & \ding{52}&{57.1}\\ 
			{} &SA + SE& 6& \ding{52} & $-$ &{58.9} \\ 
			{} &SA + SE& 6 & \ding{52} & \ding{52} &\textbf{59.8} \\
			\bottomrule[1.5pt]
		\end{tabular}
	\label{tab:ssa}
\end{table}
\begin{table}[t]
	\centering
	\caption{Ablation study of the HID module on HOD3K dataset.}
	\begin{tabular}{c|c|c|c|c|c}
		\toprule[1.5pt]
		Detectors& Input& Bands &HID & SSA  & mAP \\
		\midrule	
		{Baseline} &SA&3& $-$ & $-$ &{54.4} \\ 
		{Baseline + HID}&SA + SE&6 & \ding{52} & $-$ &{56.3}  \\
		\midrule	
		{Baseline + SSA} &SA&3 & $-$ & \ding{52}  &{54.9} \\
		{S2ADet} &SA + SE&6 & \ding{52} & \ding{52}  &\textbf{59.8}  \\
		\bottomrule[1.5pt]
	\end{tabular}
	\label{tab:hid}
\end{table}
\subsection{Ablation studies}
\emph{1) Effectiveness of different input:}
We performed a comprehensive ablation study on S2ADet to assess the impact of various input images on detection performance, as presented in Table~\ref{tab:input}. The YOLOv5 single-stream model served as the baseline. Before input, we obtained three images by decoupling hyperspectral information: SE information, pre-generation SA information, and post-generation SA information.

To evaluate the efficacy of decoupling, we fed the three aggregated images into the baseline model as input. The results indicate that the post-generation spatial aggregation image produced the best outcomes, followed by the SE information. Notably, the SA information demonstrated a 41.1\% improvement in mAP50 after generation, surpassing the SE information. The result suggests that the generation process significantly impacts the processing of spatial aggregation information. Furthermore, to validate the complementarity of spectral and spatial aggregated information, we employed different combinations of aggregated information as input to S2ADet. The results demonstrate that the combination of spectral and spatial information yielded the best performance, with a 2.6\% mAP advantage over S2Det, which exclusively uses SA information as input. These findings further confirm the effectiveness of our generated spectral and SA information.

\emph{2) Effectiveness of hyperspectral information decoupling:}
The objective of the HID module is to perform decoupling-based aggregation of spectral and spatial information in the hyperspectral image. To demonstrate the effectiveness of the HID module in S2ADet, we present a breakdown of the benefits of each internal component in Table~\ref{tab:hid}.

The results demonstrate that the standalone HID module surpasses the baseline performance by a margin of 1.9\% mAP. Furthermore, the synergistic combination of the HID and SSA modules contributes to a substantial improvement of 4.9\% mAP in the S2ADet performance. This indicates that when employed concurrently, the HID and SSA modules exhibit a complementary effect, augmenting the detector's overall efficacy. Our comprehensive experimental evaluation reveals that each constituent of the two modules enhances the HOD3K dataset performance.

\emph{3) Effectiveness of spectral-spatial aggregation module:}

The objective of the SSA module in S2ADet is to aggregate spectral and spatial information in the hyperspectral image. To assess the effectiveness of the SSA module, we present an analysis of the benefits of each internal component in Table~\ref{tab:ssa}.

The results indicate that the SSA module alone outperforms SAM by 1.8\% mAP. Moreover, the SSA module without SAM achieves a 2.6\% mAP improvement, indicating that detection performance can be enhanced by utilizing complementary information from spectral and spatial data. By employing spectral-spatial aggregation, S2ADet can achieve a 3.5\% mAP performance improvement. Hence, the SSA and SAM modules can complement each other to enhance detection performance further. Our combined experiments demonstrate that the benefits of each component of the two combinations consistently improve the detection of HOD3K.

\begin{figure*}[t]		
	\centering
	\begin{tabular}{c}
		\includegraphics[width = 1\linewidth]{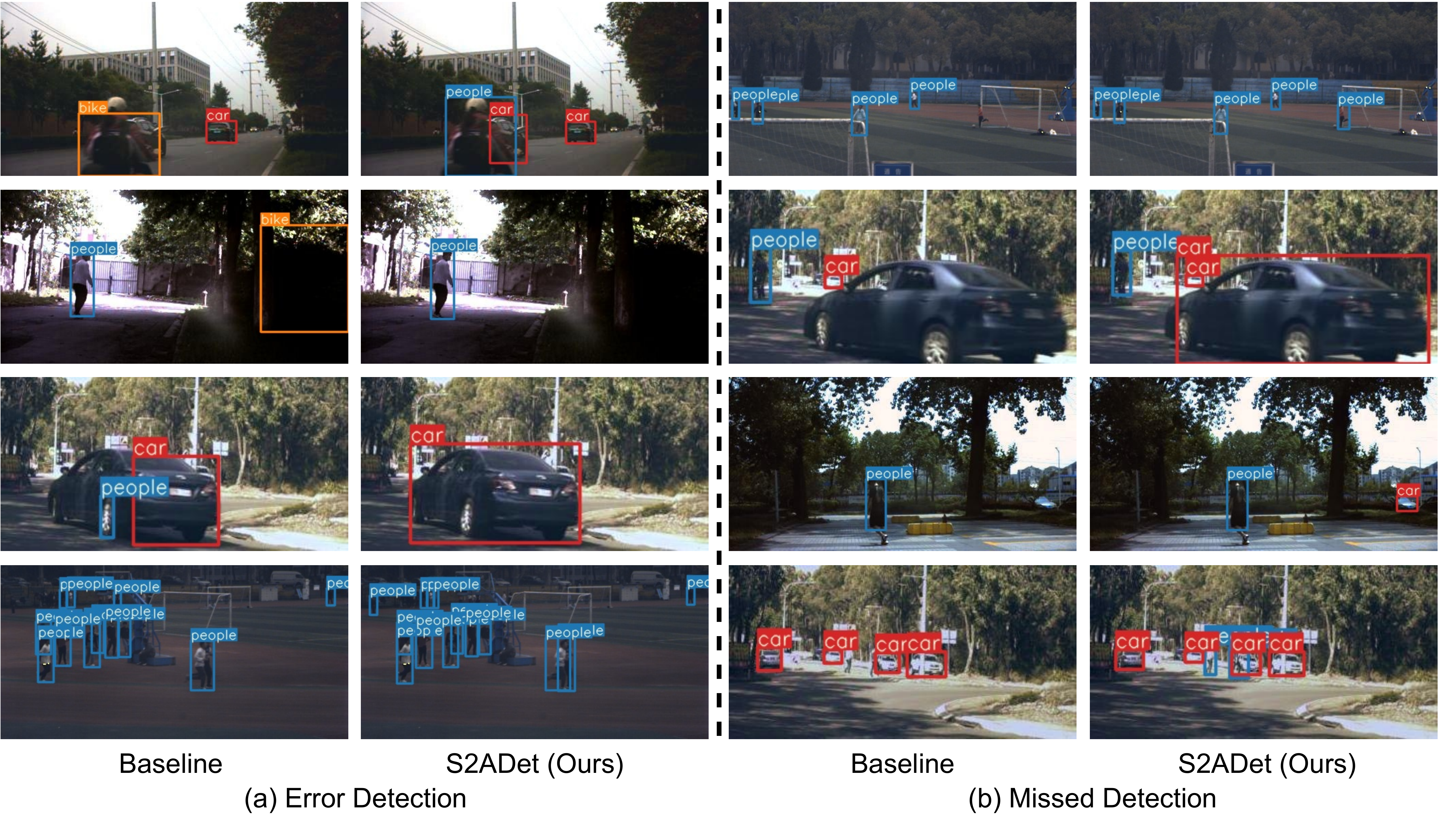}
	\end{tabular}
	\caption{Qualitative analysis of S2ADet on HOD3K. The first column in (a) and (b) shows the baseline performance, and the second column shows the performance of S2ADet. (a) The baseline model treats some redundant background information as objects. (b) S2ADet can mitigate the omission problem of the baseline model. For the analysis, we transformed the hyperspectral image into pseudo-color images for display.}
	\label{img:error}
\end{figure*}
\subsection{Qualitative analysis}

We conducted a thorough qualitative evaluation of the S2ADet algorithm on the HOD3K dataset, comparing its performance with that of the baseline model and S2ADet. We selected various scenarios to provide a comprehensive analysis, as shown in Fig.~\ref{img:error}. The first column displays the results of the baseline model, while the second column presents the performance of S2ADet.

Fig.~\ref{img:error} (a) demonstrates that S2ADet can detect objects even in the presence of occlusion by utilizing contextual information. In contrast, the baseline model struggles to identify the objects due to occlusion. Additionally, S2ADet effectively learns to differentiate between objects possessing similar features that could be easily mistaken for one another by extracting and combining spectral and spatially aggregated data, resulting in the precise identification of each object within the hyperspectral image.

Moreover, {Fig.~\ref{img:error} (b) shows that S2ADet accurately distinguishes between object and background categories based on the spectral aggregated information of the objects, mitigating the issue of small object misdetection that the baseline model encounters. The results demonstrate the effectiveness of the S2ADet algorithm in detecting objects in challenging natural scene environments.
	
The results of our qualitative evaluation indicate that the S2ADet algorithm outperforms the baseline model in detecting objects in natural scenes. By utilizing both spectral and spatial aggregated information and contextual information, S2ADet effectively addresses the limitations of the baseline model, resulting in improved object detection accuracy.
\begin{figure*}[t]		
	\centering
	\begin{tabular}{c}
		\centering
		\includegraphics[width = 1\linewidth]{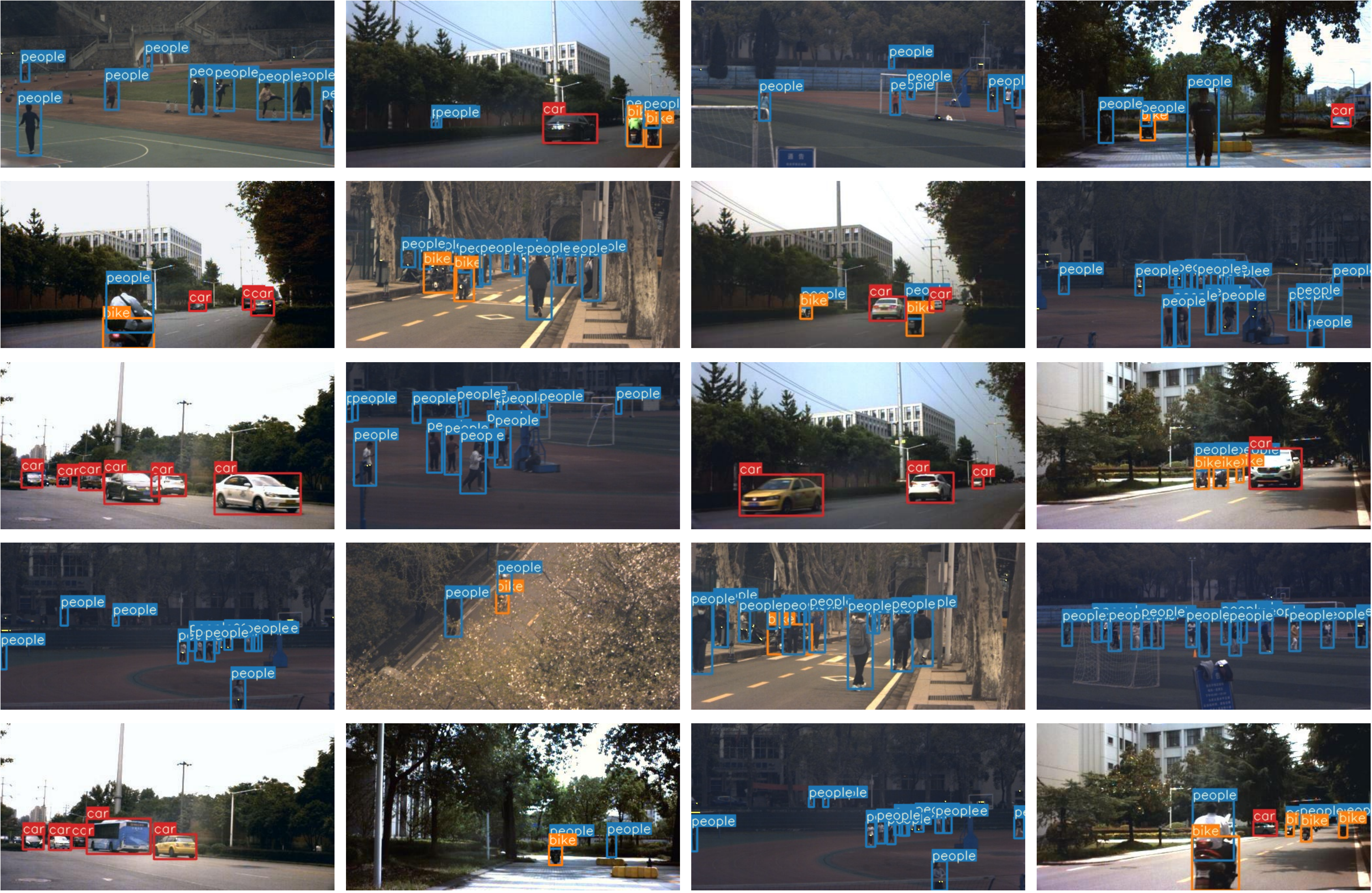}
	\end{tabular}
	\caption{Visualize some object detection results. The categories of people, bikes, and cars are represented by blue, yellow, and red boxes, respectively. We transformed the hyperspectral images into pseudo-color images for display.}
	\label{img:showtime}
\end{figure*}
\begin{figure}[t]		
	\centering
	\begin{tabular}{c}
		\centering
		\includegraphics[width = 1\linewidth]{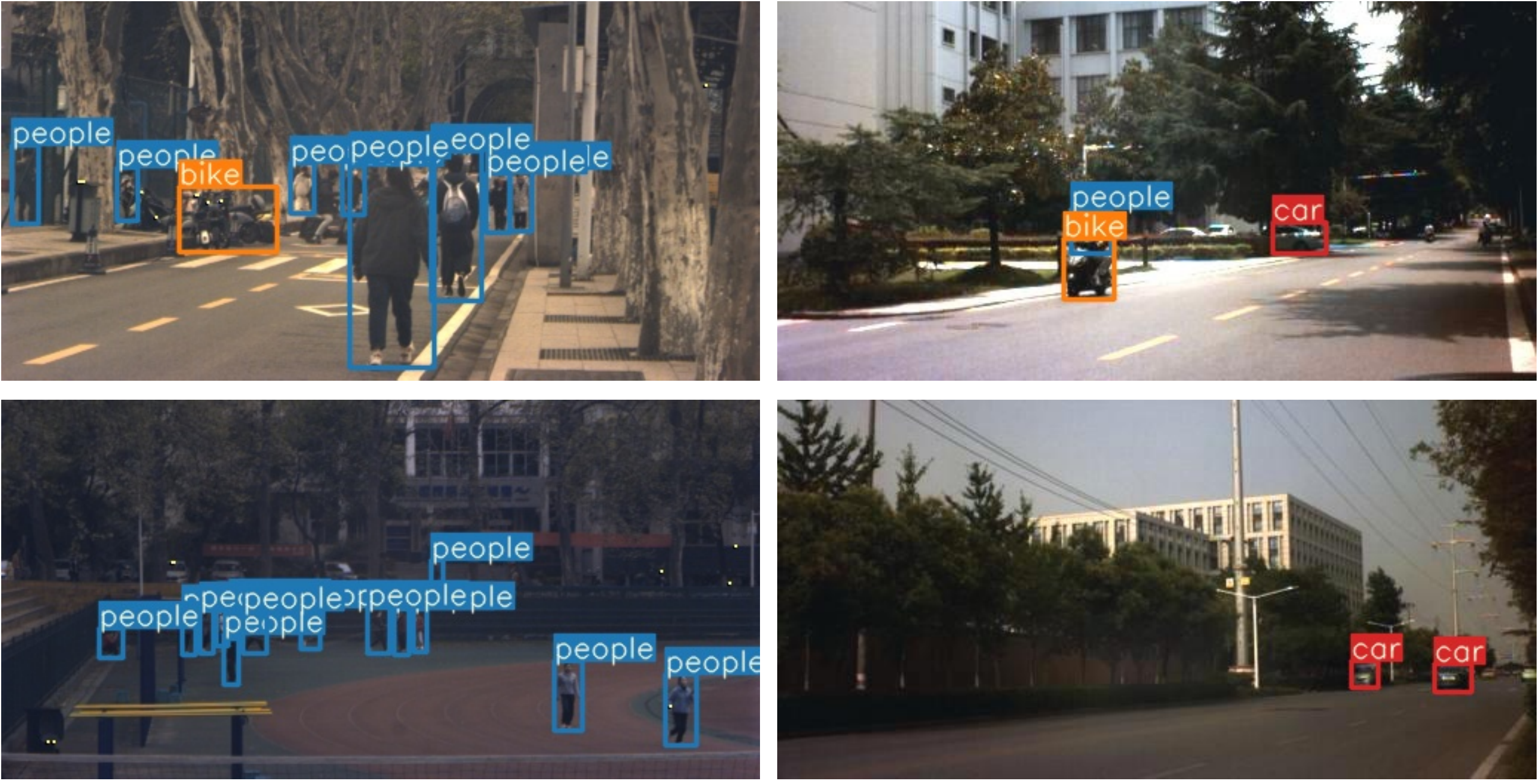}
	\end{tabular}
	\caption{Visualize some shortcoming detection results.}
	\label{img:diss}
\end{figure}
\subsection{Discussion and Visualization}
To further demonstrate the effectiveness of the S2ADet algorithm, we present additional detection results on the HOD3K dataset in Fig.~\ref{img:showtime}. The detection results showcase the algorithm's capability to accurately identify and localize objects in various natural scene environments, such as urban roads and campuses.

The high accuracy of S2ADet is evident from the clear and precise bounding boxes that accurately encompass each object. These results substantiate the efficacy of the proposed approach in hyperspectral and provide visual evidence of the algorithm's effectiveness in detecting objects in challenging natural scene environments. The visualization results demonstrate the superior performance of S2ADet over the baseline model, reinforcing the importance of considering spectral and spatial aggregated data and contextual information for object detection in the hyperspectral image. Overall, the promising performance of S2ADet on the HOD3K dataset highlights its potential to advance the object detection field in the hyperspectral image and opens up new avenues for future research.

Furthermore, we showed the classification confusion matrix for both datasets. As shown in Fig.~\ref{img:hod3k_c} (a), on the HOD-3K dataset, the pen-screen category tends to be misclassified as a background category due to the small target size of people, resulting in a classification accuracy of only 0.86. Conversely, the pen-screen category on the HOD-1 dataset exhibits an accuracy of merely 0.31, as shown in Fig.~\ref{img:hod3k_c} (b), with the majority being classified into the background category. Consequently, the algorithm warrants further improvement in the realm of fine-grained classification.


\begin{figure}[t]
  \centering
	\subfigure[]{
		\includegraphics[width=0.5\textwidth]{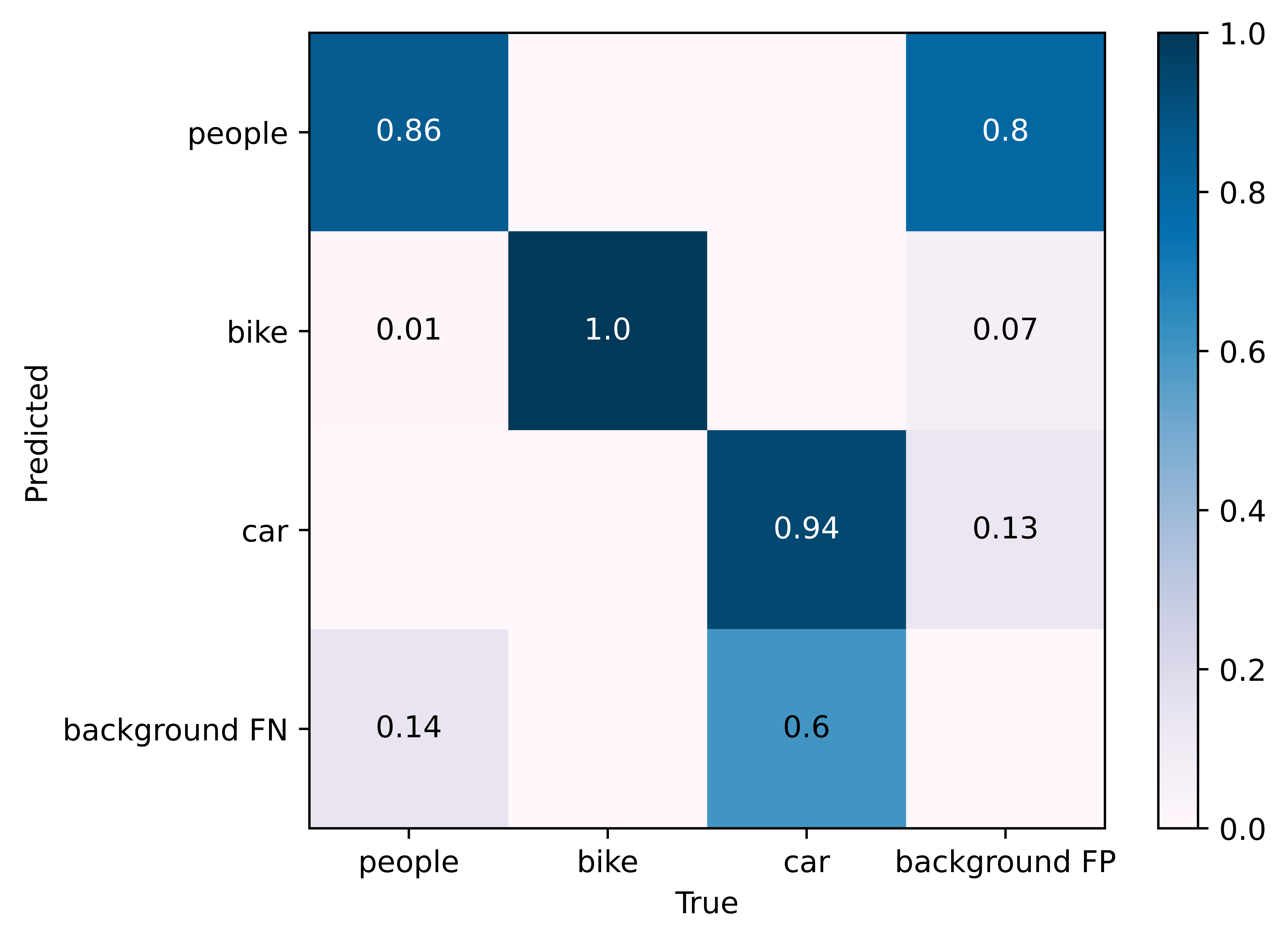}}
	\subfigure[]{
		\includegraphics[width=0.5\textwidth]{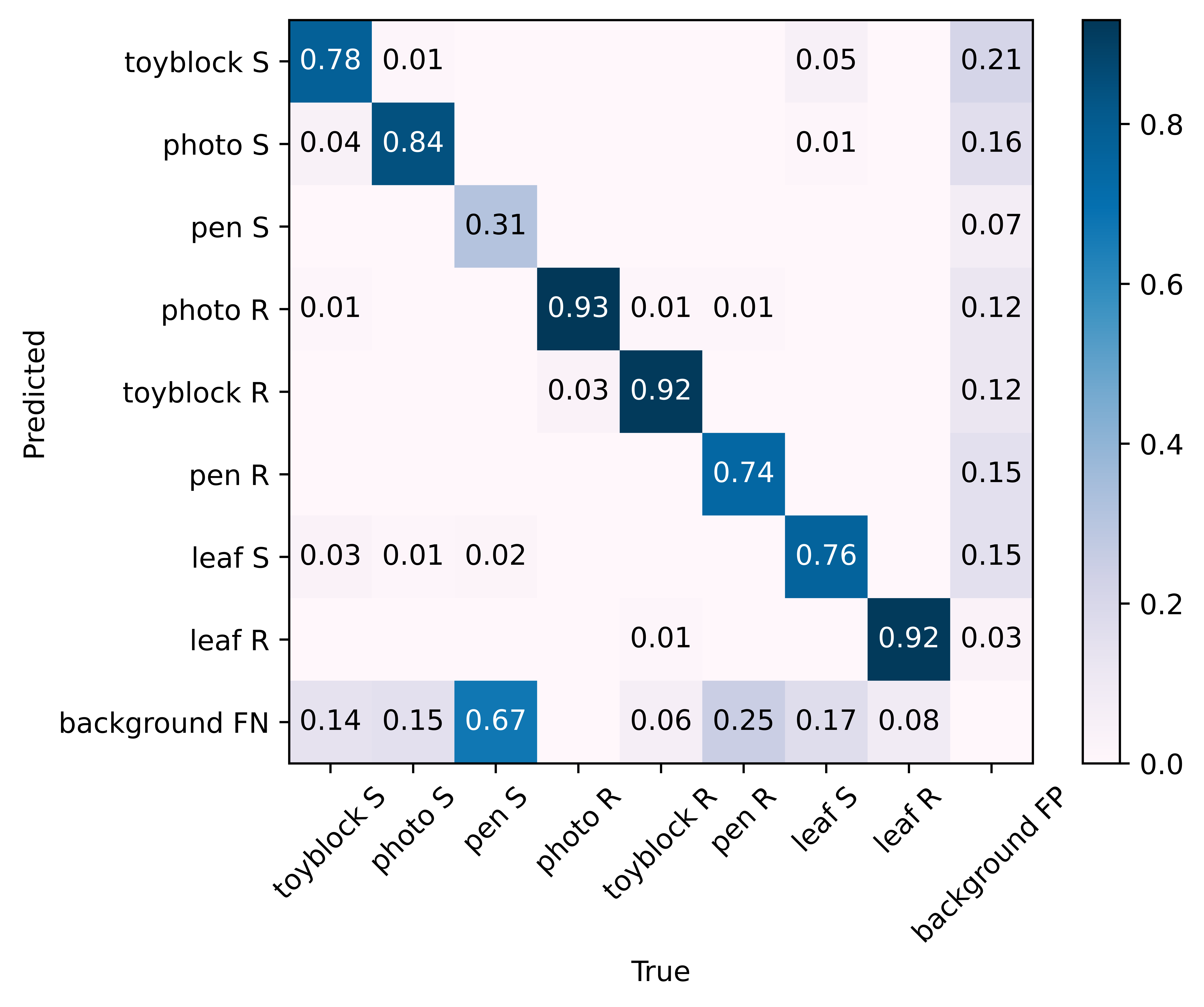}}
   \caption{Illustration of the classification confusion matrix. (a) HOD3K dataset, (b) HOD-1 dataset.}
   	\label{img:hod3k_c}
\end{figure}

\subsection{Limitations}
Although the proposed S2ADet has made significant advances in hyperspectral object detection, addressing specific challenges remains crucial for further performance enhancement. Fig.~\ref{img:diss} depicts representative failure instances from the HOD-3K dataset, where prediction errors and missed detections by S2ADet are apparent. Notably, S2ADet is challenging to predict overlapping target regions and manage small objects.

Nevertheless, the S2ADet algorithm achieved a classification accuracy exceeding 0.9 for most categories. The advancements delineated in Sections \ref{sec:sec3} and \ref{sec:sec4} attest to the proposed S2ANet's proficiency in effectively tackling the most exigent scenarios, surpassing state-of-the-art methods in hyperspectral object detection tasks.

\section{Conclusion}\label{sec:sec6}
In this paper, we propose a novel object detector S2ADet for the hyperspectral image. The proposed framework reduces hyperspectral redundancy and aggregates spectral and spatial information utilizing an HID module. A two-stream feature extraction network is then used to extract spectral and spatial features, and an SSA module is embedded in the network to mine the complementary information of spectral and spatial features using global sensory attention. Comprehensive experimental analysis verifies the effectiveness of the proposed framework and its internal modules. Our approach achieves state-of-the-art performance on two hyperspectral object detection datasets compared to other methods. The comprehensiveness of our proposed datasets and the robust applicability of our method allows our work to contribute significantly to the development of hyperspectral object detection. Moreover, we present a large-scale object detection dataset of 3242 hyperspectral images from multiple scenes. This dataset addresses the limitations of a single scene and small volume data in existing hyperspectral object detection datasets, thus facilitating the development of object detection. The HOD3K dataset is expected to advance the development of reliable and accurate hyperspectral detection systems.

\bibliographystyle{IEEEtran}
\bibliography{IEEEabrv,References}

\begin{thebibliography}{10}
\providecommand{\url}[1]{#1}
\csname url@samestyle\endcsname
\providecommand{\newblock}{\relax}
\providecommand{\bibinfo}[2]{#2}
\providecommand{\BIBentrySTDinterwordspacing}{\spaceskip=0pt\relax}
\providecommand{\BIBentryALTinterwordstretchfactor}{4}
\providecommand{\BIBentryALTinterwordspacing}{\spaceskip=\fontdimen2\font plus
\BIBentryALTinterwordstretchfactor\fontdimen3\font minus
  \fontdimen4\font\relax}
\providecommand{\BIBforeignlanguage}[2]{{%
\expandafter\ifx\csname l@#1\endcsname\relax
\typeout{** WARNING: IEEEtran.bst: No hyphenation pattern has been}%
\typeout{** loaded for the language `#1'. Using the pattern for}%
\typeout{** the default language instead.}%
\else
\language=\csname l@#1\endcsname
\fi
#2}}
\providecommand{\BIBdecl}{\relax}
\BIBdecl

\bibitem{xia2018dota}
G.-S. Xia, X.~Bai, J.~Ding, Z.~Zhu, S.~Belongie, J.~Luo, M.~Datcu, M.~Pelillo,
  and L.~Zhang, ``Dota: A large-scale dataset for object detection in aerial
  images,'' in \emph{Proceedings of the IEEE Conference on Computer Vision and
  Pattern Recognition}, 2018, pp. 3974--3983.

\bibitem{zhao2022scene}
Y.~Zhao, Y.~Zhang, Z.~Gong, and H.~Zhu, ``Scene representation in bird's-eye
  view from surrounding cameras with transformers,'' in \emph{Proceedings of
  the IEEE Conference on Computer Vision and Pattern Recognition}, 2022, pp.
  4511--4519.

\bibitem{zhang2021dynamic}
M.~Zhang, J.~Liu, Y.~Wang, Y.~Piao, S.~Yao, W.~Ji, J.~Li, H.~Lu, and Z.~Luo,
  ``Dynamic context-sensitive filtering network for video salient object
  detection,'' in \emph{Proceedings of the IEEE International Conference on
  Computer Vision}, 2021, pp. 1553--1563.

\bibitem{zhao2019object}
Z.-Q. Zhao, P.~Zheng, S.-t. Xu, and X.~Wu, ``Object detection with deep
  learning: A review,'' \emph{IEEE Transactions on Neural Networks and Learning
  Systems}, vol.~30, no.~11, pp. 3212--3232, 2019.

\bibitem{cai2022mask}
Y.~Cai, J.~Lin, X.~Hu, H.~Wang, X.~Yuan, Y.~Zhang, R.~Timofte, and L.~Van~Gool,
  ``Mask-guided spectral-wise transformer for efficient hyperspectral image
  reconstruction,'' in \emph{Proceedings of the IEEE Conference on Computer
  Vision and Pattern Recognition}, 2022, pp. 17\,502--17\,511.

\bibitem{zhang2021learning}
S.~Zhang, L.~Wang, L.~Zhang, and H.~Huang, ``Learning tensor low-rank prior for
  hyperspectral image reconstruction,'' in \emph{Proceedings of the IEEE
  Conference on Computer Vision and Pattern Recognition}, 2021, pp.
  12\,006--12\,015.

\bibitem{miao2019net}
X.~Miao, X.~Yuan, Y.~Pu, and V.~Athitsos, ``l-net: Reconstruct hyperspectral
  images from a snapshot measurement,'' in \emph{Proceedings of the IEEE
  International Conference on Computer Vision}, 2019, pp. 4059--4069.

\bibitem{mou2020nonlocal}
L.~Mou, X.~Lu, X.~Li, and X.~X. Zhu, ``Nonlocal graph convolutional networks
  for hyperspectral image classification,'' \emph{IEEE Transactions on
  Geoscience and Remote Sensing}, vol.~58, no.~12, pp. 8246--8257, 2020.

\bibitem{hou2021hyperspectral}
S.~Hou, H.~Shi, X.~Cao, X.~Zhang, and L.~Jiao, ``Hyperspectral imagery
  classification based on contrastive learning,'' \emph{IEEE Transactions on
  Geoscience and Remote Sensing}, vol.~60, pp. 1--13, 2021.

\bibitem{hong2021spectralformer}
D.~Hong, Z.~Han, J.~Yao, L.~Gao, B.~Zhang, A.~Plaza, and J.~Chanussot,
  ``Spectralformer: Rethinking hyperspectral image classification with
  transformers,'' \emph{IEEE Transactions on Geoscience and Remote Sensing},
  vol.~60, pp. 1--15, 2021.

\bibitem{rui2021learning}
X.~Rui, X.~Cao, Q.~Xie, Z.~Yue, Q.~Zhao, and D.~Meng, ``Learning an explicit
  weighting scheme for adapting complex hsi noise,'' in \emph{Proceedings of
  the IEEE Conference on Computer Vision and Pattern Recognition}, 2021, pp.
  6739--6748.

\bibitem{chang2020effective}
C.-I. Chang, ``An effective evaluation tool for hyperspectral target detection:
  3d receiver operating characteristic curve analysis,'' \emph{IEEE
  Transactions on Geoscience and Remote Sensing}, vol.~59, no.~6, pp.
  5131--5153, 2020.

\bibitem{dong2021asymmetric}
Y.~Dong, W.~Shi, B.~Du, X.~Hu, and L.~Zhang, ``Asymmetric weighted logistic
  metric learning for hyperspectral target detection,'' \emph{IEEE Transactions
  on Cybernetics}, vol.~52, no.~10, pp. 11\,093--11\,106, 2021.

\bibitem{shang2020target}
X.~Shang, M.~Song, Y.~Wang, C.~Yu, H.~Yu, F.~Li, and C.-I. Chang,
  ``Target-constrained interference-minimized band selection for hyperspectral
  target detection,'' \emph{IEEE Transactions on Geoscience and Remote
  Sensing}, vol.~59, no.~7, pp. 6044--6064, 2020.

\bibitem{chang2021hyperspectral}
C.-I. Chang, ``Hyperspectral anomaly detection: A dual theory of hyperspectral
  target detection,'' \emph{IEEE Transactions on Geoscience and Remote
  Sensing}, vol.~60, pp. 1--20, 2021.

\bibitem{sun2021constrained}
X.~Sun, H.~Zhang, F.~Xu, Y.~Zhu, and X.~Fu, ``Constrained-target band selection
  with subspace partition for hyperspectral target detection,'' \emph{IEEE
  Journal of Selected Topics in Applied Earth Observations and Remote Sensing},
  vol.~14, pp. 9147--9161, 2021.

\bibitem{liu2016tensor}
Y.~Liu, G.~Gao, and Y.~Gu, ``Tensor matched subspace detector for hyperspectral
  target detection,'' \emph{IEEE Transactions on Geoscience and Remote
  Sensing}, vol.~55, no.~4, pp. 1967--1974, 2016.

\bibitem{yan2021object}
L.~Yan, M.~Zhao, X.~Wang, Y.~Zhang, and J.~Chen, ``Object detection in
  hyperspectral images,'' \emph{IEEE Signal Processing Letters}, vol.~28, pp.
  508--512, 2021.

\bibitem{neuenschwander1998mapping}
A.~L. Neuenschwander, M.~M. Crawford, and M.~J. Provancha, ``Mapping of coastal
  wetlands via hyperspectral aviris data,'' in \emph{IGARSS'98. Sensing and
  Managing the Environment. 1998 IEEE International Geoscience and Remote
  Sensing. Symposium Proceedings.(Cat. No. 98CH36174)}, vol.~1.\hskip 1em plus
  0.5em minus 0.4em\relax IEEE, 1998, pp. 189--191.

\bibitem{resmini1997mineral}
R.~Resmini, M.~Kappus, W.~Aldrich, J.~Harsanyi, and M.~Anderson, ``Mineral
  mapping with hyperspectral digital imagery collection experiment (hydice)
  sensor data at cuprite, nevada, usa,'' \emph{International Journal of Remote
  Sensing}, vol.~18, no.~7, pp. 1553--1570, 1997.

\bibitem{sun2020scalability}
P.~Sun, H.~Kretzschmar, X.~Dotiwalla, A.~Chouard, V.~Patnaik, P.~Tsui, J.~Guo,
  Y.~Zhou, Y.~Chai, B.~Caine \emph{et~al.}, ``Scalability in perception for
  autonomous driving: Waymo open dataset,'' in \emph{Proceedings of the
  IEEE/CVF conference on computer vision and pattern recognition}, 2020, pp.
  2446--2454.

\bibitem{zavrtanik2021draem}
V.~Zavrtanik, M.~Kristan, and D.~Sko{\v{c}}aj, ``Draem-a discriminatively
  trained reconstruction embedding for surface anomaly detection,'' in
  \emph{Proceedings of the IEEE/CVF International Conference on Computer
  Vision}, 2021, pp. 8330--8339.

\bibitem{girshick2015fast}
R.~Girshick, ``Fast r-cnn,'' in \emph{Proceedings of the IEEE International
  Conference on Computer Vision}, 2015, pp. 1440--1448.

\bibitem{he2017mask}
K.~He, G.~Gkioxari, P.~Doll{\'a}r, and R.~Girshick, ``Mask r-cnn,'' in
  \emph{Proceedings of the IEEE international conference on computer vision},
  2017, pp. 2961--2969.

\bibitem{zhang2021vit}
Z.~Zhang, X.~Lu, G.~Cao, Y.~Yang, L.~Jiao, and F.~Liu, ``Vit-yolo:
  Transformer-based yolo for object detection,'' in \emph{Proceedings of the
  IEEE/CVF international conference on computer vision}, 2021, pp. 2799--2808.

\bibitem{liu2016ssd}
W.~Liu, D.~Anguelov, D.~Erhan, C.~Szegedy, S.~Reed, C.-Y. Fu, and A.~C. Berg,
  ``Ssd: Single shot multibox detector,'' in \emph{Computer Vision--ECCV 2016:
  14th European Conference, Amsterdam, The Netherlands, October 11--14, 2016,
  Proceedings, Part I 14}.\hskip 1em plus 0.5em minus 0.4em\relax Springer,
  2016, pp. 21--37.

\bibitem{girshick2014rich}
R.~Girshick, J.~Donahue, T.~Darrell, and J.~Malik, ``Rich feature hierarchies
  for accurate object detection and semantic segmentation,'' in
  \emph{Proceedings of the IEEE conference on computer vision and pattern
  recognition}, 2014, pp. 580--587.

\bibitem{ren2015faster}
S.~Ren, K.~He, R.~Girshick, and J.~Sun, ``Faster r-cnn: Towards real-time
  object detection with region proposal networks,'' \emph{Advances in neural
  information processing systems}, vol.~28, 2015.

\bibitem{liu2020deep}
L.~Liu, W.~Ouyang, X.~Wang, P.~Fieguth, J.~Chen, X.~Liu, and
  M.~Pietik{\"a}inen, ``Deep learning for generic object detection: A survey,''
  \emph{International journal of computer vision}, vol. 128, pp. 261--318,
  2020.

\bibitem{redmon2017yolo9000}
J.~Redmon and A.~Farhadi, ``Yolo9000: better, faster, stronger,'' in
  \emph{Proceedings of the IEEE conference on computer vision and pattern
  recognition}, 2017, pp. 7263--7271.

\bibitem{feng2021tood}
C.~Feng, Y.~Zhong, Y.~Gao, M.~R. Scott, and W.~Huang, ``Tood: Task-aligned
  one-stage object detection,'' in \emph{2021 IEEE/CVF International Conference
  on Computer Vision (ICCV)}.\hskip 1em plus 0.5em minus 0.4em\relax IEEE
  Computer Society, 2021, pp. 3490--3499.

\bibitem{tian2019fcos}
Z.~Tian, C.~Shen, H.~Chen, and T.~He, ``Fcos: Fully convolutional one-stage
  object detection,'' in \emph{Proceedings of the IEEE International Conference
  on Computer Vision}, 2019, pp. 9627--9636.

\bibitem{carion2020end}
N.~Carion, F.~Massa, G.~Synnaeve, N.~Usunier, A.~Kirillov, and S.~Zagoruyko,
  ``End-to-end object detection with transformers,'' in \emph{Computer
  Vision--ECCV 2020: 16th European Conference, Glasgow, UK, August 23--28,
  2020, Proceedings, Part I 16}.\hskip 1em plus 0.5em minus 0.4em\relax
  Springer, 2020, pp. 213--229.

\bibitem{hang2020classification}
R.~Hang, Z.~Li, P.~Ghamisi, D.~Hong, G.~Xia, and Q.~Liu, ``Classification of
  hyperspectral and lidar data using coupled cnns,'' \emph{IEEE Transactions on
  Geoscience and Remote Sensing}, vol.~58, no.~7, pp. 4939--4950, 2020.

\bibitem{bai2020class}
J.~Bai, A.~Yuan, Z.~Xiao, H.~Zhou, D.~Wang, H.~Jiang, and L.~Jiao, ``Class
  incremental learning with few-shots based on linear programming for
  hyperspectral image classification,'' \emph{IEEE Transactions on
  Cybernetics}, vol.~52, no.~6, pp. 5474--5485, 2020.

\bibitem{liu2020few}
S.~Liu, Q.~Shi, and L.~Zhang, ``Few-shot hyperspectral image classification
  with unknown classes using multitask deep learning,'' \emph{IEEE Transactions
  on Geoscience and Remote Sensing}, vol.~59, no.~6, pp. 5085--5102, 2020.

\bibitem{zhu2019binary}
D.~Zhu, B.~Du, and L.~Zhang, ``Binary-class collaborative representation for
  target detection in hyperspectral images,'' \emph{IEEE Geoscience and Remote
  Sensing Letters}, vol.~16, no.~7, pp. 1100--1104, 2019.

\bibitem{sellami2022deep}
A.~Sellami and S.~Tabbone, ``Deep neural networks-based relevant latent
  representation learning for hyperspectral image classification,''
  \emph{Pattern Recognition}, vol. 121, p. 108224, 2022.

\bibitem{wang2019hyperspectral}
L.~Wang, C.~Sun, Y.~Fu, M.~H. Kim, and H.~Huang, ``Hyperspectral image
  reconstruction using a deep spatial-spectral prior,'' in \emph{Proceedings of
  the IEEE Conference on Computer Vision and Pattern Recognition}, 2019, pp.
  8032--8041.

\bibitem{li2023lightweight}
Z.~Li, C.~Tang, X.~Liu, W.~Zhang, J.~Dou, L.~Wang, and A.~Y. Zomaya,
  ``Lightweight remote sensing change detection with progressive feature
  aggregation and supervised attention,'' \emph{IEEE Transactions on Geoscience
  and Remote Sensing}, vol.~61, pp. 1--12, 2023.

\bibitem{li2021mapping}
Q.~Li, F.~K.~K. Wong, and T.~Fung, ``Mapping multi-layered mangroves from
  multispectral, hyperspectral, and lidar data,'' \emph{Remote Sensing of
  Environment}, vol. 258, p. 112403, 2021.

\bibitem{farmonov2023crop}
N.~Farmonov, K.~Amankulova, J.~Szatm{\'a}ri, A.~Sharifi, D.~Abbasi-Moghadam,
  S.~M.~M. Nejad, and L.~Mucsi, ``Crop type classification by desis
  hyperspectral imagery and machine learning algorithms,'' \emph{IEEE Journal
  of Selected Topics in Applied Earth Observations and Remote Sensing},
  vol.~16, pp. 1576--1588, 2023.

\bibitem{liu2021anchor}
Z.~Liu, X.~Wang, M.~Shu, G.~Li, C.~Sun, Z.~Liu, and Y.~Zhong, ``An anchor-free
  siamese target tracking network for hyperspectral video,'' in \emph{2021 11th
  Workshop on Hyperspectral Imaging and Signal Processing: Evolution in Remote
  Sensing (WHISPERS)}.\hskip 1em plus 0.5em minus 0.4em\relax IEEE, 2021, pp.
  1--5.

\bibitem{koz2019ground}
A.~Koz, ``Ground-based hyperspectral image surveillance systems for explosive
  detection: Part i—state of the art and challenges,'' \emph{IEEE Journal of
  Selected Topics in Applied Earth Observations and Remote Sensing}, vol.~12,
  no.~12, pp. 4746--4753, 2019.

\bibitem{fang2023hyperspectral}
L.~Fang, Y.~Jiang, Y.~Yan, J.~Yue, and Y.~Deng, ``Hyperspectral image instance
  segmentation using spectral--spatial feature pyramid network,'' \emph{IEEE
  Transactions on Geoscience and Remote Sensing}, vol.~61, pp. 1--13, 2023.

\bibitem{vali2020deep}
A.~Vali, S.~Comai, and M.~Matteucci, ``Deep learning for land use and land
  cover classification based on hyperspectral and multispectral earth
  observation data: A review,'' \emph{Remote Sensing}, vol.~12, no.~15, p.
  2495, 2020.

\bibitem{hong2020learning}
D.~Hong, J.~Chanussot, N.~Yokoya, J.~Kang, and X.~X. Zhu, ``Learning-shared
  cross-modality representation using multispectral-lidar and hyperspectral
  data,'' \emph{IEEE Geoscience and Remote Sensing Letters}, vol.~17, no.~8,
  pp. 1470--1474, 2020.

\bibitem{fang2023towards}
L.~Fang, Y.~Yan, J.~Yue, and Y.~Deng, ``Towards the vectorization of
  hyperspectral imagery,'' \emph{IEEE Transactions on Geoscience and Remote
  Sensing}, 2023.

\bibitem{wu2023querying}
L.~Wu, L.~Fang, X.~He, M.~He, J.~Ma, and Z.~Zhong, ``Querying labeled for
  unlabeled: Cross-image semantic consistency guided semi-supervised semantic
  segmentation,'' \emph{IEEE Transactions on Pattern Analysis and Machine
  Intelligence}, 2023.

\bibitem{li2019deep}
S.~Li, W.~Song, L.~Fang, Y.~Chen, P.~Ghamisi, and J.~A. Benediktsson, ``Deep
  learning for hyperspectral image classification: An overview,'' \emph{IEEE
  Transactions on Geoscience and Remote Sensing}, vol.~57, no.~9, pp.
  6690--6709, 2019.

\bibitem{wang2020hyperspectral}
Q.~Wang, F.~Zhang, and X.~Li, ``Hyperspectral band selection via optimal
  neighborhood reconstruction,'' \emph{IEEE Transactions on Geoscience and
  Remote Sensing}, vol.~58, no.~12, pp. 8465--8476, 2020.

\bibitem{chen2021hyperspectral}
H.~Chen, F.~Miao, Y.~Chen, Y.~Xiong, and T.~Chen, ``A hyperspectral image
  classification method using multifeature vectors and optimized kelm,''
  \emph{IEEE Journal of Selected Topics in Applied Earth Observations and
  Remote Sensing}, vol.~14, pp. 2781--2795, 2021.

\bibitem{chang2021orthogonal}
C.-I. Chang, H.~Cao, and M.~Song, ``Orthogonal subspace projection target
  detector for hyperspectral anomaly detection,'' \emph{IEEE Journal of
  Selected Topics in Applied Earth Observations and Remote Sensing}, vol.~14,
  pp. 4915--4932, 2021.

\bibitem{yu2017convolutional}
S.~Yu, S.~Jia, and C.~Xu, ``Convolutional neural networks for hyperspectral
  image classification,'' \emph{Neurocomputing}, vol. 219, pp. 88--98, 2017.

\bibitem{xiong2020material}
F.~Xiong, J.~Zhou, and Y.~Qian, ``Material based object tracking in
  hyperspectral videos,'' \emph{IEEE Transactions on Image Processing},
  vol.~29, pp. 3719--3733, 2020.

\bibitem{torralba2010labelme}
A.~Torralba, B.~C. Russell, and J.~Yuen, ``Labelme: Online image annotation and
  applications,'' \emph{Proceedings of the IEEE}, vol.~98, no.~8, pp.
  1467--1484, 2010.

\bibitem{lin2017feature}
T.-Y. Lin, P.~Doll{\'a}r, R.~Girshick, K.~He, B.~Hariharan, and S.~Belongie,
  ``Feature pyramid networks for object detection,'' in \emph{Proceedings of
  the IEEE conference on computer vision and pattern recognition}, 2017, pp.
  2117--2125.

\bibitem{abdi2010principal}
H.~Abdi and L.~J. Williams, ``Principal component analysis,'' \emph{Wiley
  interdisciplinary reviews: computational statistics}, vol.~2, no.~4, pp.
  433--459, 2010.

\bibitem{guo2022cmt}
J.~Guo, K.~Han, H.~Wu, Y.~Tang, X.~Chen, Y.~Wang, and C.~Xu, ``Cmt:
  Convolutional neural networks meet vision transformers,'' in
  \emph{Proceedings of the IEEE Conference on Computer Vision and Pattern
  Recognition}, 2022, pp. 12\,175--12\,185.

\bibitem{woo2018cbam}
S.~Woo, J.~Park, J.-Y. Lee, and I.~S. Kweon, ``Cbam: Convolutional block
  attention module,'' in \emph{Proceedings of the European Conference on
  Computer Vision}, 2018, pp. 3--19.

\bibitem{pang2019libra}
J.~Pang, K.~Chen, J.~Shi, H.~Feng, W.~Ouyang, and D.~Lin, ``Libra r-cnn:
  Towards balanced learning for object detection,'' in \emph{Proceedings of the
  IEEE Conference on Computer Vision and Pattern Recognition}, 2019, pp.
  821--830.

\bibitem{chen2021you}
Q.~Chen, Y.~Wang, T.~Yang, X.~Zhang, J.~Cheng, and J.~Sun, ``You only look
  one-level feature,'' in \emph{Proceedings of the IEEE/CVF conference on
  computer vision and pattern recognition}, 2021, pp. 13\,039--13\,048.

\bibitem{zhu2020deformable}
X.~Zhu, W.~Su, L.~Lu, B.~Li, X.~Wang, and J.~Dai, ``Deformable detr: Deformable
  transformers for end-to-end object detection,'' \emph{arXiv preprint
  arXiv:2010.04159}, 2020.

\bibitem{lin2017focal}
T.-Y. Lin, P.~Goyal, R.~Girshick, K.~He, and P.~Doll{\'a}r, ``Focal loss for
  dense object detection,'' in \emph{Proceedings of the IEEE International
  Conference on Computer Vision}, 2017, pp. 2980--2988.

\bibitem{wu2020rethinking}
Y.~Wu, Y.~Chen, L.~Yuan, Z.~Liu, L.~Wang, H.~Li, and Y.~Fu, ``Rethinking
  classification and localization for object detection,'' in \emph{Proceedings
  of the IEEE Conference on Computer Vision and Pattern Recognition}, 2020, pp.
  10\,186--10\,195.

\bibitem{kong2020foveabox}
T.~Kong, F.~Sun, H.~Liu, Y.~Jiang, L.~Li, and J.~Shi, ``Foveabox: Beyound
  anchor-based object detection,'' \emph{IEEE Transactions on Image
  Processing}, vol.~29, pp. 7389--7398, 2020.

\end{thebibliography}

\end{document}